\documentclass{article}

\usepackage{arxiv}
\usepackage[backend=biber,style=nature]{biblatex}
\RequirePackage{graphicx,xcolor,ae,fancyvrb}
\DefineVerbatimEnvironment{Sinput}{Verbatim}{fontshape=sl}
\DefineVerbatimEnvironment{Soutput}{Verbatim}{}
\DefineVerbatimEnvironment{Scode}{Verbatim}{fontshape=sl}

\DefineVerbatimEnvironment{Code}{Verbatim}{}
\DefineVerbatimEnvironment{CodeInput}{Verbatim}{fontshape=sl}
\DefineVerbatimEnvironment{CodeOutput}{Verbatim}{}
\newenvironment{CodeChunk}{}{}

\usepackage[utf8]{inputenc} 
\usepackage[english]{babel}
\usepackage[T1]{fontenc}    
\usepackage{hyperref}       
\usepackage{url}            
\usepackage{booktabs}       
\usepackage{amsfonts}       
\usepackage{nicefrac}       
\usepackage{microtype}      
\usepackage{lipsum}		    
\usepackage{graphicx}
\usepackage{doi}
\usepackage{subcaption}
\usepackage{orcidlink}

\usepackage{caption}
\usepackage{float}
\usepackage[section]{placeins}
\usepackage{amsmath} 
\usepackage{listings} 
\usepackage{pbox}
\usepackage{tabularx}

\graphicspath{{./img/}}

\title{\pkg{IM}: An \proglang{R}-Package for Computation of Image Moments and Moment Invariants}
\date{April 25, 2014}	

\author{Allison Irvine and M. Murat Dundar\\
	Dept. of Computer \& Information Science\\
	Indiana University -- Purdue University, Indianapolis\\
	723 W. Michigan Street\\
	Indianapolis, IN 46202\\
	\And
	Tan Dang\\
	Dept. of Mathematics\\
	Purdue University\\
	150 N. University Street\\
	West Lafayette, IN 47907\\
	\AND
	Bartek Rajwa~\orcidlink{0000-0001-7540-8236}\\
	Bindley Bioscience Center\\
	Purdue University\\
	1203 W. State Street\\
	West Lafayette, IN 47907\\
	E-mail: \email{brajwa@purdue.edu}\\
	URL: \url{http://web.ics.purdue.edu/~brajwa}\\
}

\renewcommand{\headeright}{Technical Report}
\renewcommand{\undertitle}{Technical Report}
\renewcommand{\shorttitle}{IM Package}

\hypersetup{
pdftitle={IM: An R Package for Computation of Image Moments and Moment Invariants},
pdfsubject={cs.CV},
pdfauthor={Allison Irvine,  Elias Tan Dang, M. Murat Dundar, Bartek Rajwa},
pdfkeywords={Zernike, Pseudo-Zernike, Fourier, Hahn, Chebyshev, Legendre, Gegenbauer, polynomials, moments, moment invariants, image analysis},
}

\begin{document}
\maketitle

\begin{abstract}
Moment invariants are well-established and effective shape descriptors for image classification. In this report, we introduce a package for \proglang{R}-language, named \pkg{IM}, that implements the calculation of moments for images and allows the reconstruction of images from moments within an object-oriented framework. Several types of moments may be computed using the \pkg{IM} library, including discrete and continuous Chebyshev, Gegenbauer, Legendre, Krawtchouk, dual Hahn, generalized pseudo-Zernike, Fourier-Mellin, and radial harmonic Fourier moments. In addition, custom bivariate types of moments can be calculated using combinations of two different types of polynomials. A method of polar transformation of pixel coordinates is used to provide an approximate invariance to rotation for moments that are orthogonal over a rectangle. The different types of polynomials used to calculate moments are discussed in this report, as well as comparisons of reconstruction and running time. Examples of image classification using image moments are provided.
\end{abstract}

\keywords{Zernike, Pseudo-Zernike, Fourier, Hahn, Chebyshev, Legendre, Gegenbauer, polynomials, moments, moment invariants, image analysis, R-language}

\section{Introduction}
Continuously developing technologies for image acquisition are resulting in an abundance of digital images collected for various fields of scientific research. In many cases, the answers to vital research questions may be found by quantifying similarities and differences between images. Although many machine-learning algorithms exist for efficient and robust data classification, without effective feature extraction, their usefulness remains limited. Moment invariants have been widely used in image recognition  \cite{Flusser1994,Flusser2009,Mukundan2004,Persson2001} since 1962, when Hu published his well-known paper describing the use of seven moment invariants for planar geometric figures \cite{Hu1962}. Moments are values that describe the distribution of a function and are commonly used to describe the distribution of mass in rigid-body mechanics \cite{Flusser2009,Hu1962}. Some moments commonly encountered are descriptive statistics such as mean, variance, skewness, and kurtosis. 

A grayscale image may be viewed as a distribution of pixel intensities in two dimensions. Moment invariants, derived from moments, are shape descriptors that are insensitive to variations such as position (translation), size (scaling), and orientation (rotation) \cite{Hu1962,Persson2001}. Many types of moment invariants exist and have been used for various image recognition and classification tasks. Among the applications of moment invariants have been character recognition \cite{Papakostas2010}, object recognition \cite{Mukundan1995,Mukundan2001,Mukundan2004}, image watermarking \cite{Venkataramana2007}, aircraft identification \cite{Dudani1977}, leaf-shape analysis \cite{Persson2001}, and identification of bacteria \cite{Bayraktar2006,Forero2004,Rajwa2010}. In \cite{Bayraktar2006}, Zernike moment invariants were employed to extract features from circular light-scatter patterns formed by bacterial colonies analyzed with a laser scatterometer. These features were processed with LDA and PCA feature-reduction and hierarchical-clustering algorithms to identify different strains of {\it Listeria}. In \cite{Rajwa2010}, the Zernike moment invariants were paired with a class discovery system allowing for the detection of emerging pathogens. Despite the popularity of moment invariants in image analysis, there is no widely available library or package allowing students, educators, and researchers to experiment with various moment-based feature-extraction techniques.

Herein we report the development of an R package named \pkg{IM}. This package allows the computation of several types of moments and moment invariants from images in an object-oriented framework. Generalized pseudo-Zernike, Chebyshev, Hahn, Krawtchouk, Legendre, Gegenbauer, Fourier Mellin, Fourier Chebyshev, radial harmonic Fourier, and bivariate moment types are all available in the package. There are many similarities in the computation of moments of varying types, but the main distinction between the reported and implemented variants is the polynomial function used to calculate moments. These moments describe shapes and textures in digital images up to varying detail orders. Different types of moments have particular advantages and disadvantages with respect to image degradation, detail, reconstruction, and complexity. Some additional functionality is present in the package for plotting polynomials and moments, as well as computing steps required prior to moment calculation, such as centroid calculation and polar coordinate transformation. 

In subsequent sections, we briefly define and describe the image moments, the moment invariants, and the various types of polynomials employed for calculation. We present the methods used to calculate the moments and some simple examples of image classification and reconstruction. A brief comparison of the moments available in the \pkg{IM} package is also reported.

\subsection{Moments}
Moments of an image may be defined as a weighted sum of the pixels over a coordinate space \cite{Sheng1994}, where the moment weighting kernel is defined by the type of polynomials used \cite{Flusser2009,Mukundan1998}. The order of the moment weighting kernel determines the amount of detail that will be contained in the moments \cite{Mukundan1998}. The first few orders of geometric moments of pixel values with respect to their location in an image describe general features such as the total image area, the centroid location, and the orientation \cite{Flusser2009,Mukundan1998,Sheng1994}.

If a grayscale image is treated as a two-dimensional distribution of pixel intensities, it may be represented by the function 
\begin{equation} \label{eq:img} \begin{array}{cc} f(x,y), & \mathrm{for }\, x \in 1,\dots,N_x, y \in 1,\dots,N_y \\ \end{array} \end{equation}
where $x$ and $y$ are the coordinates of the pixels within the image space, $f(x,y)$ is the intensity of the pixel located at $(x,y)$, and $N_x$ and $N_y$ are the horizontal and vertical dimensions of the image. The moments of an image, $M_{pq}$, are generally calculated by taking the integral of the product of the image and a weighting kernel. 
\begin{equation} \label{eq:conv}M_{pq} = \int_{N_x} \int_{N_y} K_{pq}(x,y)f(x,y)dxdy \end{equation}
The dimensions of the image define the space under which the integral will be calculated. The moment weighting kernel $K_{pq}$ is a function of $x$ and $y$ and depends on constants $p$ and $q$, where $(p+q)$ is the order of the kernel. Often the order defines the degree of a polynomial in the weighting kernel function, but may have a different role in the function depending on the type of polynomial used. The simplest type of moments are geometric moments, whose weighting kernel $K_{pq}$ is defined by the function
\begin{equation} \label{eq:geometric} K_{pq}(x,y) = x^{p}y^{q} \end{equation}

Geometric moments of orders 0 to 3 describe general shape information such as the total area, centroid location, and orientation of the image \cite{Mukundan1998}. Similar to polynomial regression, using a higher-order polynomial to approximate a set of values results in a more exact approximation, while lower-order polynomial results in a rougher approximation \cite{Flusser2009, Mukundan1998}. However, a more exact approximation of values is sensitive to noise and leads to over-fitting in classification problems \cite{Bayraktar2006}. Therefore, while higher-order moments may be used to produce a better reconstruction of an image, lower-order moments may be more beneficial for an application where moment invariants are used for the classification of many vaguely similar images. Some types of moments, such as discrete Chebyshev moments, may be used to produce a perfect image reconstruction if moments up to an order equal to the image dimensions are used.

Two main types of moments are implemented in the \pkg{IM} package. The first type, which we refer to as \emph{orthogonal} moments, is calculated over the euclidean coordinate space. The second type, referred to as \emph{complex} moments, are calculated over the polar coordinate space. Although both types are technically orthogonal, the moments explicitly referred to as orthogonal moments are computed using real-valued polynomials that are orthogonal over a rectangle. In contrast, complex moments are calculated using complex polynomials that are orthogonal over the unit disk \cite{Flusser2009, Bayraktar2006, Mukundan1998}. The orthogonal moments characterize independent features of an image, therefore minimizing information redundancy \cite{Bayraktar2006, Mukundan1995, Mukundan1998, Papakostas2010}.

\subsection{Moment invariants}
Moment invariants are functions of moments that may be used to describe an image in a way that is insensitive to certain deformations or transformations \cite{Dudani1977, Flusser1994, Flusser2009, Hu1962, Mukundan1998, Ping2002, Rui1999, Sheng1994}. The problems of image classification and object recognition often involve identifying several images of the same object that have each undergone some transformation that makes them technically nonidentical \cite{Bayraktar2006, Dudani1977, Flusser1994, Hu1962, Persson2001, Rui1999} even though they may convey identical information. The most common types of deformations are translation \emph{t}, rotation \emph{R}, and scaling \emph{s}. The observed image $\hat{I}$ is often a function of the original image $I$ and the degrading variables.
\begin{equation} \label{eq:deform}\hat{I}=sRI + t \end{equation}
A function of an image is invariant with respect to a particular type of deformation if its value remains the same whether or not the image has undergone such a transformation \cite{Flusser2009, Hu1962, Mukundan1998}. Invariance with respect to translation may be achieved by shifting the coordinates of an object so that its centroid is at the origin of the coordinate space before calculating moments \cite{Flusser2009}. Moment invariants with respect to scaling may be derived by normalizing the values of the image moments \cite{Flusser2009} or by normalizing the image before calculating moments \cite{Papakostas2010}. Complex moments are preferred for calculating moment invariants with respect to rotation because they are defined over polar coordinates \cite{Bayraktar2006, Flusser2009, Mukundan1995, Mukundan1998, Ping2002, Sheng1994}. The moment invariants with respect to rotation are simply the magnitudes of these complex moments \cite{Flusser2009}.

\section{Orthogonal moments}
Orthogonal moments are calculated using a kernel that satisfies a relation of orthogonality or weighted orthogonality. These types of moments are also more robust to random noise and result in a reconstruction that minimizes the mean-square error when using a finite set of moments, since the moments are coordinates of the image function $f(x,y)$ in the polynomial basis. Orthogonal polynomials can be calculated quickly, and without information, loss using recurrence relations \cite{Flusser2009}. This is an advantage of using orthogonal polynomials instead of geometric polynomials as defined by Equation \ref{eq:geometric}, which can result in a large range of values, creating problems due to limitations of numeric precision in software \cite{Flusser2009, Mukundan1995}. For a two-dimensional image, the kernel is a product of two one-dimensional polynomials. This allows one to construct kernels that are a product of two different types of polynomials; a technique also implemented in the \pkg{IM} package \cite{Zhu2012}. Classical orthogonal polynomials may be continuous or discrete \cite{Zhu2012}. While orthogonal moments are optimal for image reconstruction \cite{Mukundan2004}, they are not often used for image compression since better methods, such as wavelet-based compression, are available. 

An expression encountered in many of the formal definitions of the polynomials is the hypergeometric function, defined as
\begin{equation}\label{eq:hypergeometric}_2F_1(a,b;c;z) = \sum_{n=0}^\infty \frac{(a)_n (b)_n}{(c)_n}\frac{z^n}{n!}\end{equation}
This leads us to define the Pochhammer symbol as
\begin{equation} \label{eq:pochhammer}
	{\left( q \right)_n} = \left\{ {\begin{array}{*{20}{c}}
			1&{{\rm{if}}}&{n = 0}\\
			{q\left( {q + 1} \right) \ldots \left( {q + n - 1} \right)}&{{\rm{if}}}&{n > 0}
	\end{array}} \right.
\end{equation}

In some cases the generalized hypergeometric series is more useful. The generalized hypergeometric series is
\begin{equation}\label{eq:genhypergeometric}_pF_q(a_1,\dots,a_p;b_1,\dots,b_q;z) = \sum_{n=0}^{\infty} \frac{(a_1)_n\dots(a_p)_n}{(b_1)_n\dots(b_q)_n}\frac{z^n}{n!}\end{equation}

\subsection{Continuous orthogonal moments}
Continuous moments $M_{pq}$ of orders $p$ and $q$ in the $x$ and $y$ directions are calculated using an integral, as in Equation \ref{eq:conv}, but must be approximated using summations in software \cite{Flusser2009, Zhu2012}. Continuous orthogonal polynomials satisfy the condition of orthogonality only within the range $[-1,1]$. Therefore the pixel coordinates of the image must be scaled between -1 and 1 in both the $x$ and $y$ directions \cite{Flusser2009, Zhu2012}. If the image is not square, scaling the pixel coordinates deforms the image resulting in a poor reconstruction from the moments.

Continuous orthogonal moments $M_{pq}$ are generally calculated by: 
\begin{equation}\label{contMoments}M_{pq}=\int^1_{-1} \int^1_{-1}K_{pq}(x,y)f(x,y)\,dx\,dy\end{equation}
where the weighting kernel $K_{pq}$ is the product of two polynomials.
\begin{equation}\label{kernel}K_{pq}(x,y)=k_{1p}(x)k_{2q}(y)\end{equation}
For a real discretized image this formula is approximated by the following:
\begin{equation}\label{contMomentsSum}M_{pq}=\sum_{x=-1}^{1}\sum_{y=-1}^{1} k_{1p}(x)k_{2q}(y)f(x,y)\end{equation}
$k_{1p}$ is a polynomial of order $p$ and $k_{2p}$ is a polynomial of order $q$. The distinction between the two vectors is made using the subscripts $1$ and $2$ to illustrate that each vector may be a different type of polynomial. The entire matrix of moments, $M$, containing the values $M_{pq}$ for all $p$ and $q$ is calculated in the \pkg{IM} package in matrix form using the following equation:
\begin{equation}\label{matMoments} M=k_{1}Fk_{2}\end{equation}
In \ref{matMoments}, $F$ is the entire $(N \times M)$ image matrix and $k_{1}$ and $k_{2}$ are matrices of polynomials where rows correspond to order and columns correspond to $x$ and $y$ pixel coordinates, respectively. Each pixel $\hat{f}(x,y)$ of the reconstructed image from the moments is calculated by \cite{Flusser2009,Zhu2012} 
\begin{equation}\label{contReconstructSum}\hat{f}(x,y) = \sum_{p=0}^{\infty}\sum_{q=0}^{\infty}K_{pq}(x,y)M_{pq}\end{equation}
An arbitrary maximum order of moments may be used to reconstruct an image. Recall, however, that lower-order moments capture only gross features of the image while higher-order moments capture more detail. Therefore, using a higher order of moments will yield a better reconstruction. The entire reconstructed image $\hat{F}$ in the \pkg{IM} package is implemented using the matrix form of Equation \ref{contReconstructSum}.
\begin{equation}\label{matReconstruct}\hat{F} = k_{1}Mk_{2}\end{equation}
The polynomials described in the following sections are normalized to overcome numerical precision issues due to a high range of polynomial values \cite{Mukundan2001,Mukundan2004}. In general, normalization is performed by multiplying the polynomials by 
\begin{equation}\label{scaling}\sqrt{\frac{w(x)}{\rho(p)}}\end{equation}
where $w(x)$ is a weighting function of some kind and $\rho(p)$ is the squared norm. The squared norm is derived from the relation of orthogonality \cite{Flusser2009}. If the weighting function or squared norm is not provided in the following descriptions of the polynomials, it is equal to one.

\subsection{Continuous Chebyshev polynomials} 
\begin{figure}[H]
	\begin{subfigure}[H]{0.49\textwidth}
		\centering
		\includegraphics[width=\textwidth]{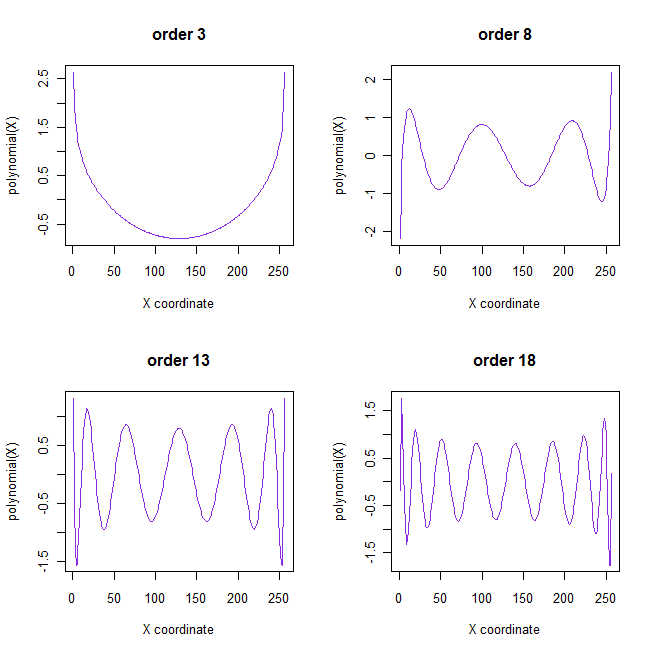}
	\end{subfigure}
	\begin{subfigure}[H]{0.49\textwidth}
		\centering
		\includegraphics[width=\textwidth]{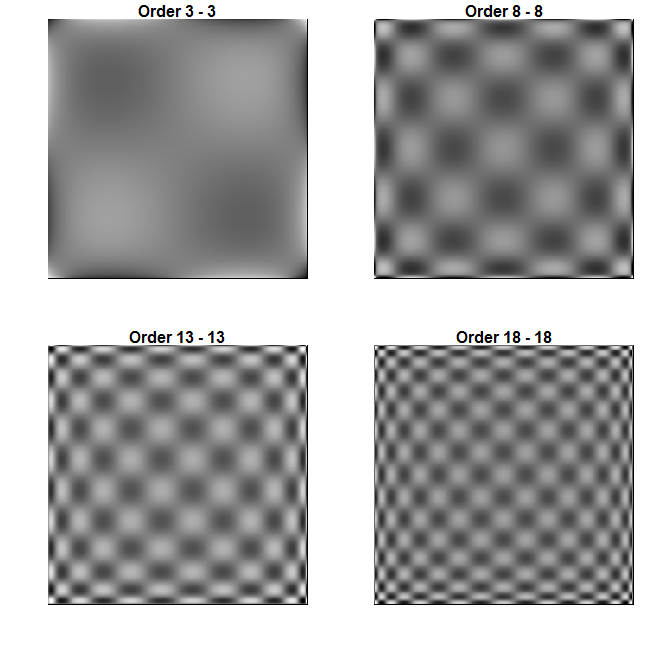}
	\end{subfigure}
	\caption{Continuous Chebyshev polynomials of orders 3, 8, 13, and 18 (left) and Chebyshev kernels, created by the combination of two Chebyshev polynomials of equal order (right).}
	\label{fig:cont-chebyshev}
\end{figure}

Continuous Chebyshev polynomials (Fig. \ref{fig:cont-chebyshev}) $k_p(x)$ of the first kind with order $p$ are formally defined as \cite{Flusser2009,Zhu2012}
\begin{equation}\label{eq:contchebypoly}{k_p}\left( x \right){ = _2}{F_1}\left( { - p,p;\frac{1}{2};\frac{{1 - x}}{2}} \right).\end{equation}
With the weighting function defined as
\begin{equation}\label{eq:contchebyweight}w(x)= (1-x^2)^{-0.5},\end{equation}
normalized Chebyshev polynomials are given by the following equation:
\begin{equation}\label{eq:contchebynormpoly}\hat{k}_p(x)= k_p(x)\sqrt{\frac{2}{\pi}}(1-x^2)^{0.25}.\end{equation}
Normalized Chebyshev polynomials of the first kind follow the recurrence relation
\begin{equation}\label{eq:contchebyrecur}k_{p+1}(x)= 2xk_{p}(x) - k_{p-1}(x),\end{equation}
with 
\[{k_0}\left( x \right) = 1,\quad {k_1}\left( x \right) = x\]
The \pkg{IM} package uses this recurrence relation to calculate Chebyshev polynomials.

\subsection{Legendre polynomials} 
\begin{figure}[H]
	\begin{subfigure}[H]{0.49\textwidth}
		\centering
		\includegraphics[width=\textwidth]{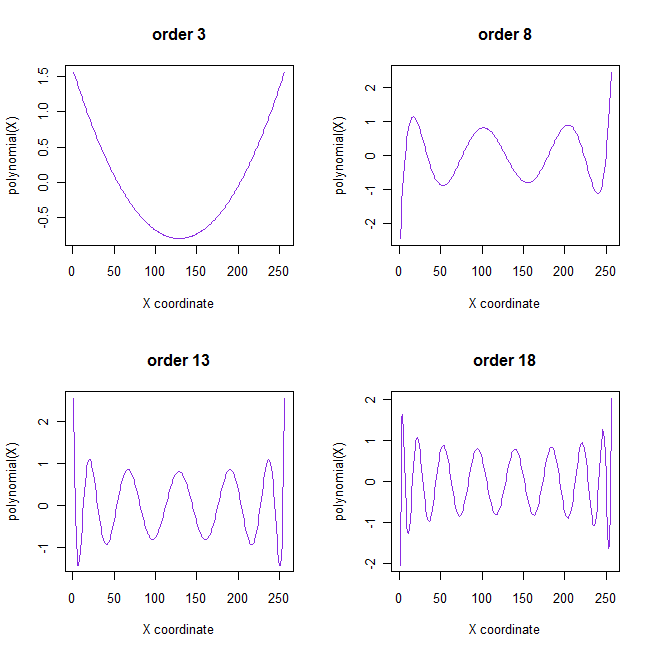}
	\end{subfigure}
	\quad
	\begin{subfigure}[H]{0.49\textwidth}
		\centering
		\includegraphics[width=\textwidth]{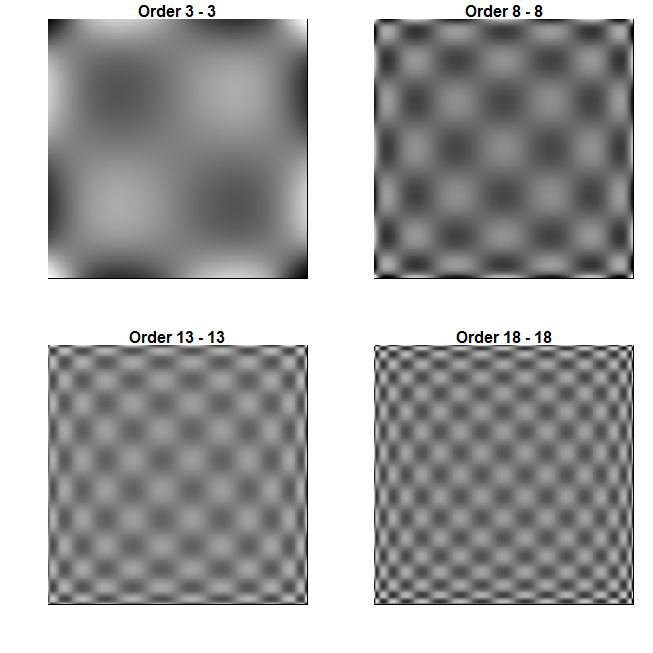}
	\end{subfigure}
	\caption{Legendre polynomials of orders 3, 8, 13, and 18 (left) and Legendre kernels, created by the combination of two Legendre polynomials of equal order (right).}
	\label{fig:legendre}
\end{figure}

Legendre polynomials (Fig. \ref{fig:legendre}) of order $p$ are defined as \cite{Flusser2009,Mukundan1995,Zhu2012}  
\begin{equation}\label{eq:legendpoly}k_p(x)= {}_2 F_1\left( -p,p+1;1;\frac{1-x}{2} \right)\end{equation}
The squared norm is given by 
\begin{equation}\label{eq:legendnorm}\frac{2}{2p+1}.\end{equation}
Normalized Legendre polynomials are then defined as 
\begin{equation}\label{eq:legendpolynorm}\hat{k}_p(x)= k_p(x)\sqrt{\frac{2p+1}{2}}.\end{equation}
Normalized Legendre polynomials are computed in the \pkg{IM} package by the recurrence formula
\begin{equation}\label{eq:legendrecur}k_{p+1}(x)= \frac{1}{p+2}\left((2p+1)xk_p(x)-pk_{p-1}\right),\end{equation}
with 
\[ k_0(x)=1, k_1(x)=x \] 

\subsection{Gegenbauer polynomials} 
\begin{figure}[H]
	\begin{subfigure}[H]{0.49\textwidth}
		\centering
		\includegraphics[width=\textwidth]{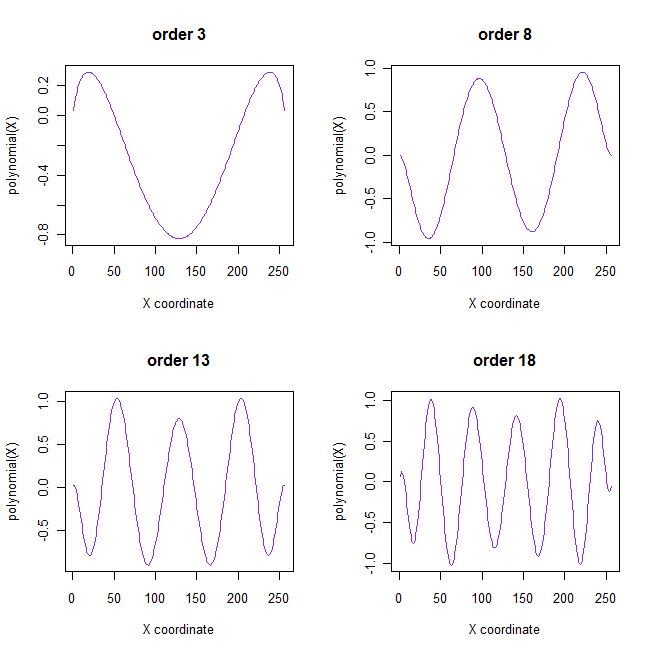}
	\end{subfigure}
	\quad
	\begin{subfigure}[H]{0.49\textwidth}
		\centering
		\includegraphics[width=\textwidth]{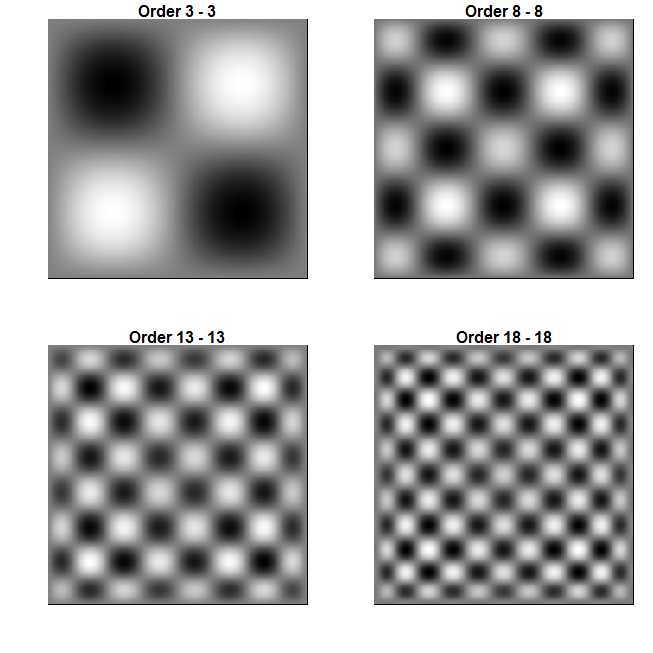}
	\end{subfigure}
	\caption{Gegenbauer polynomials with $\alpha=2$ of orders 3, 8, 13, and 18 (left) and Gegenbauer kernels, created by the combination of two Gegenbauer polynomials with $\alpha=2$ of equal order (right).}
	\label{fig:gegenbauer}
\end{figure}

Gegenbauer polynomials (Fig. \ref{fig:gegenbauer}) of order $p$ are defined as \cite{Zhu2012}
\begin{equation}\label{eq:gegenpoly}k_p^{(\alpha)}(x)= \frac{(2\alpha)_p}{p!} {}_2 F_1 \left( -p,2\alpha+1+p; \alpha+\frac{1}{2};\frac{1-x}{2} \right) \end{equation} 

This type of polynomial requires an additional parameter, $\alpha$. This parameter scales the polynomial values and is subject to the constraints
\[\alpha  >  - \frac{1}{2}\quad {\rm{and}}\quad \alpha \not  = 0\]

A larger value of $\alpha$ increases the polynomial values. If $\alpha$ is equal to 1, the resulting polynomials are equivalent to continuous Chebyshev polynomials. The weighting function $w(x)$ is given by
\begin{equation}\label{eq:gegenweight}w(x)= (1-x^2)^{\alpha -0.5}\end{equation}
and the squared norm is defined as 
\begin{equation}\label{eq:gegennorm}\rho_p(\alpha)= \frac{2\pi \Gamma(p+2\alpha)}{2^{2\alpha}p!(p+\alpha)(\Gamma(\alpha))^2}\end{equation}
Normalized Gegenbauer polynomials are then calculated by 
\begin{equation}\label{eq:gegenpolynorm}\hat{k}_p^{(\alpha)}= k_p^{(\alpha)}\sqrt{\frac{w(x)}{\rho_p(\alpha)}}\end{equation}
The recurrence relation used to compute normalized Gegenbauer polynomials is 
\begin{equation}\label{eq:gegenrecur}k_{p+1}^{(\alpha)}(x)= \frac{2p+\alpha}{p+\alpha}x k_p^{(\alpha)}(x) - \frac{p+2\alpha-1}{p+1}k_{p-1}^{(\alpha)}(x)\end{equation}
with 
\[k_0^{(\alpha )}(x) = 1,\quad k_1^{(\alpha )}(x) = 2\alpha x\]

\subsection{Discrete orthogonal moments}
Discrete orthogonal moments $M_{pq}$ of an image are generally calculated using the following equation:
\begin{equation}\label{eq:discretemoments}M_{pq}=\sum_{x=1}^{N}\sum_{y=1}^{M} k_{1p}(x)k_{2q}(y)f(x,y).\end{equation}
$N$ and $M$ are the dimensions of the image in the $x$ and $y$ directions. Discrete orthogonal polynomials are defined in the original image coordinate space and do not require any numerical approximations in software implementation \cite{Mukundan2001, Mukundan2004}. Therefore the implemented polynomials satisfy the orthogonality property and produce a better reconstruction than approximations of continuously defined polynomials \cite{Mukundan2001, Mukundan2004}. As a result, numerical errors resulting from an approximation of an integral are not propagated in the recurrence relations used to calculate the polynomials \cite{Mukundan2004}. The matrix implementations in the \pkg{IM} package for moment calculation, Equation \ref{matMoments}, and reconstruction, Equation \ref{matReconstruct}, are the same for both discrete and continuous orthogonal moments.

\subsection{Discrete Chebyshev polynomials} 
\begin{figure}[H]
	\begin{subfigure}[H]{0.49\textwidth}
		\centering
		\includegraphics[width=\textwidth]{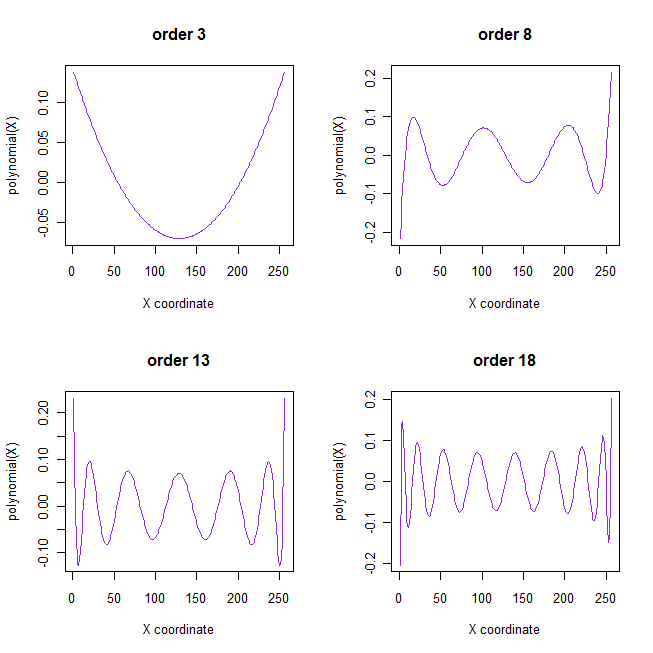}
	\end{subfigure}
	\quad
	\begin{subfigure}[H]{0.49\textwidth}
		\centering
		\includegraphics[width=\textwidth]{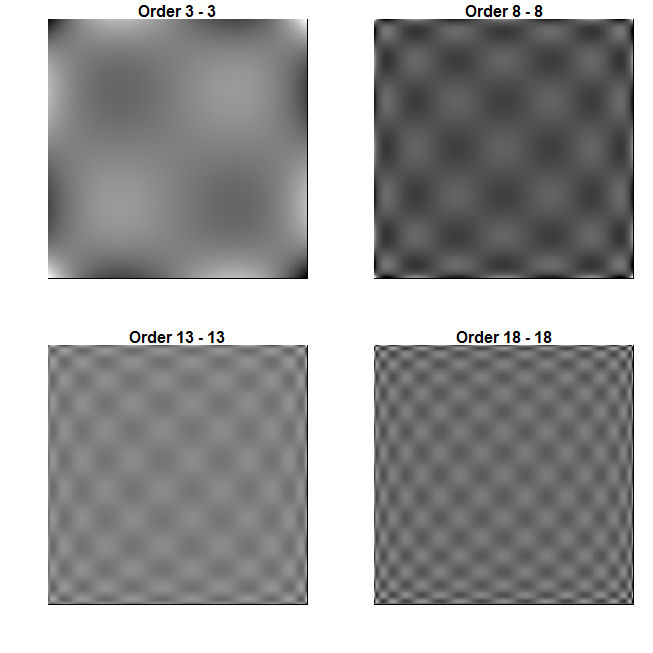}
	\end{subfigure}
	\caption{Discrete Chebyshev polynomials of orders 3, 8, 13, and 18 (left) and discrete Chebyshev kernels, created by the combination of two discrete Chebyshev polynomials of equal order (right).}
	\label{fig:dist-chebyshev}
\end{figure}

Discrete Chebyshev polynomials (Fig. \ref{fig:dist-chebyshev}) of order $p$ are defined as \cite{Mukundan2001,Mukundan2004}

\begin{equation}\label{eq:discchebypoly}
	\begin{array}{c}
		{k_p}\left( {x,N} \right) = {\left( {1 - N} \right)_p} \times {}_3{F_2}\left( { - p, - x,1 + p;1,1 - N;1} \right)\\
		{\rm{for}}\;p,x = 0,1, \ldots ,N - 1
	\end{array}
\end{equation}

The squared norm is given by
\begin{equation}\label{eq:discchebynorm}\rho(p,N)= (2p)! \binom{N+1}{2p+1}.\end{equation}
Normalized discrete Chebyshev polynomials are defined as 
\begin{equation}\label{eq:discchebynormpoly}\hat{k}_p(x,N)= k_p(x;N)\sqrt{\frac{2p+1}{N(N^2-1^2)(N^2-2^2)\dots(N^2-p^2)}}\end{equation}
The recurrence relation used to compute normalized discrete Chebyshev polynomials is 
\begin{equation}\label{eq:discchebyrecur}\hat{k}_p(x; N)= \frac{1}{x(N-x)}(A\hat{k}_p(x-1;N)+B\hat{k}_p(x-2;N))\end{equation}
with
\[
\begin{split}
	A&= -p(p+1)-(2x-1)(x-N-1)-x\\
	B&= (x-1)(x-N-1)
\end{split}
\]
and
\[
\begin{split}
	\hat{k}_p(0; N)&= (1-N)_p \times \sqrt{\frac{2p+1}{N(N^2-1^2)(N^2-2^2)\dots(N^2-p^2)}}\\
	\hat{k}_p(1;N)&= \hat{k}_p(0; N) \left( 1+ \frac{p(p+1)}{1-N} \right)
\end{split}
\]

\subsection{Krawtchouk polynomials} 
\begin{figure}[H]
	\begin{subfigure}[H]{0.49\textwidth}
		\centering
		\includegraphics[width=\textwidth]{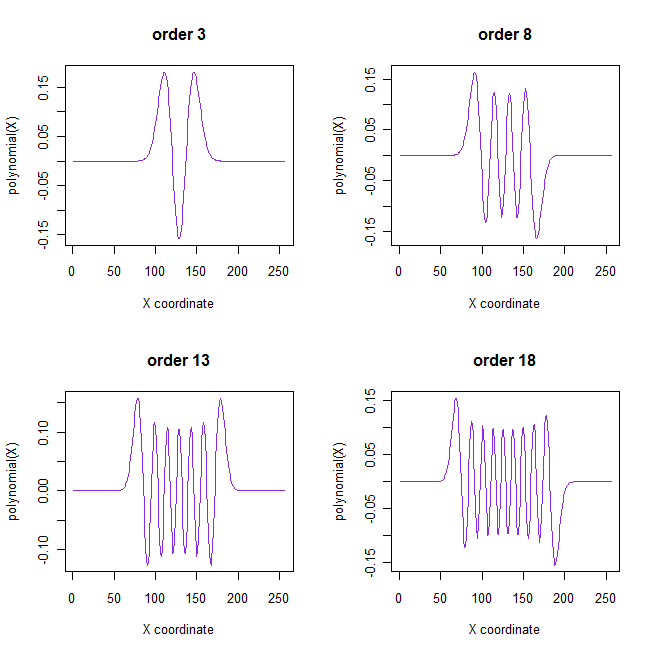}
	\end{subfigure}
	\begin{subfigure}[H]{0.49\textwidth}
		\centering
		\includegraphics[width=\textwidth]{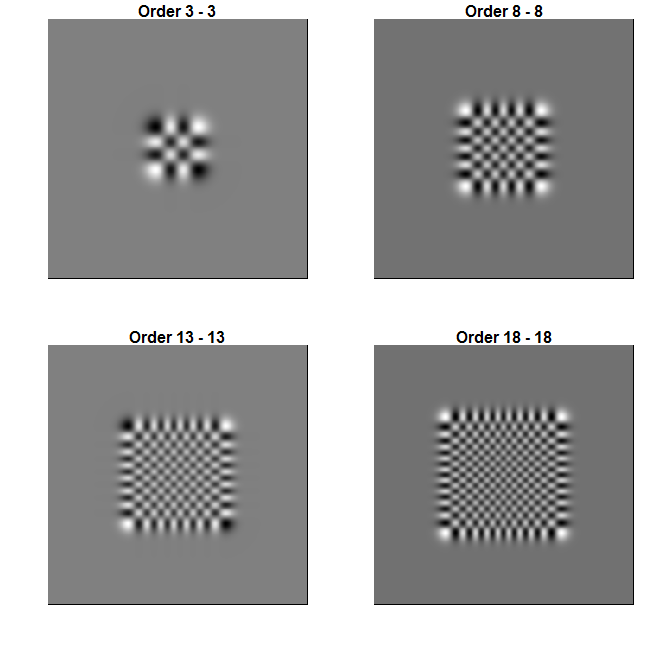}
	\end{subfigure}
	\caption{Krawtchouk polynomials with $\alpha=0.5$ of orders 3, 8, 13, and 18 (left) and Krawtchouk kernels, created by the combination of two Krawtchouk polynomials of equal order (right).}
	\label{fig:krawtchouk}
\end{figure}

Krawtchouk polynomials (Fig. \ref{fig:krawtchouk}) of order $p$ are defined as \cite{Papakostas2010,Venkataramana2007}

\begin{equation}\label{eq:krawtpoly}
	\begin{split}{l}
		{k_p}(x;\alpha ,N){ = _2}{F_1}\left( {p, - x;1 - N;\frac{1}{\alpha }} \right)\\
		{\rm{for}}\quad p,x = 0,1,2, \ldots ,N - 1
	\end{split}
\end{equation}

This type of polynomial requires an additional parameter, $\alpha$. This parameter localizes the polynomial values between 0 and $N$, the dimension of the image. For example, if $\alpha = 0$, the polynomial is centered at the first pixel coordinate. If $\alpha = \frac{1}{2}$, the center of the polynomial is shifted to the center of the image. This parameter is subject to the constraint $\alpha \in (0,1)$

The effect of the parameter $\alpha$ on the location of the polynomial indicates that Krawtchouk moments are effective local descriptors \cite{Papakostas2010}. The weighting function is defined as  
\begin{equation}\label{eq:krawtweight}w(x) =\binom{N-1}{x} ^x(1-\alpha)^{N-1-x} \end{equation}
The squared norm is given by
\begin{equation}\label{eq:krawtnorm} \rho(p)= (-1)^p \left( \frac{1-\alpha}{\alpha} \right)^p \frac{p!}{(1-N)_p} \end{equation}
Normalized Krawtchouk polynomials are then calculated using the following equation:
\begin{equation}\label{eq:krawtnormpoly}\hat{k}_p(x;\alpha,N)= k_p(x;\alpha, N)\sqrt{\frac{w(x)}{\rho(p)}}\end{equation}
The recurrence relation used to calculate normalized Krawtchouk polynomials is
\begin{equation}\label{krawtrecur}\hat{k}_{p+1}(x;\alpha,N)= \frac{1}{\alpha(N-1-p)}(A \times\hat{k}_p(x;\alpha,N)- B\times\hat{k}_{p-1}(x;\alpha,N))\end{equation}

where
\[
\begin{split}
	A&= \sqrt{\frac{\alpha(N-1-p)}{(1-\alpha)(p+1)}}((N-1)\alpha-2p\alpha+p-x)\\
	B&= \sqrt{\frac{\alpha^2 (N-1-p)(N-p)}{(1-\alpha)^2(p+1)p}}(p(1-\alpha))
\end{split}
\]
and 
\[
\begin{split}
	\hat{k}_0(x;\alpha,N)&= \sqrt{w(x)}\\
	\hat{k}_1(x;\alpha,N)&= \left(1-\frac{x}{\alpha(N-1)}\right)\left(\sqrt{\frac{\alpha(N-1)}{1-\alpha}}\right)\sqrt{w(x)}
\end{split}
\]

\subsection{Dual Hahn polynomials}
\begin{figure}[H]
	\centering
	\begin{subfigure}[H]{0.49\textwidth}
		\centering
		\includegraphics[width=\textwidth]{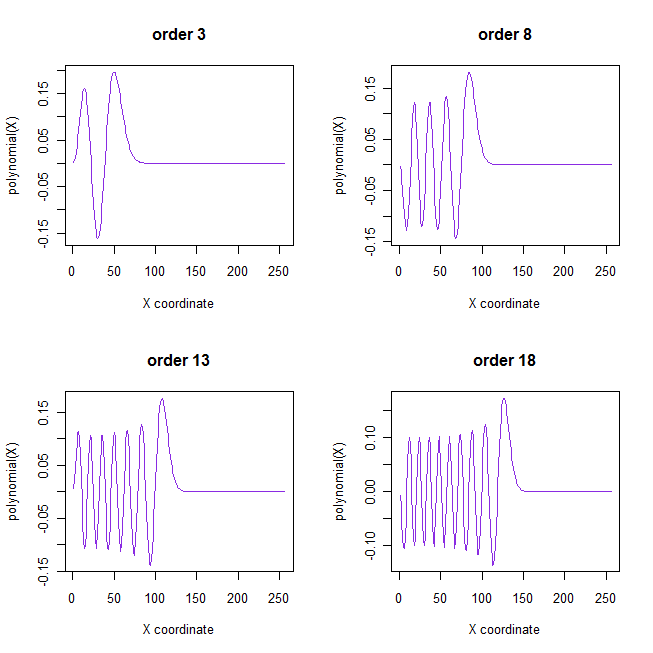}
	\end{subfigure}
	\begin{subfigure}[H]{0.49\textwidth}
		\centering
		\includegraphics[width=\textwidth]{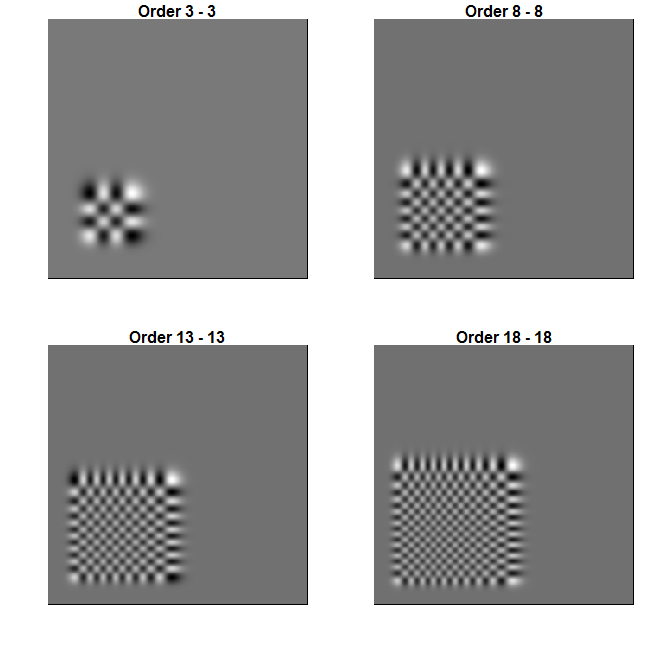}
	\end{subfigure}
	\caption{Dual Hahn polynomials with $a=10$ and $c=10$ of orders 3, 8, 13, and 18 (left) and Dual Hahn kernels, created by the combination of two dual Hahn polynomials with $a=10$ and $c=10$ of equal order (right).}
	\label{fig:hahn}
\end{figure}

Dual Hahn polynomials (Fig. \ref{fig:hahn}) are defined on a nonuniform lattice $x(s) = s(s+1)$, where $x$ is the distance of a sample from the origin and $s$ is the order of a sample \cite{Flusser2009}. Dual Hahn polynomials of order $p$ are defined as 
\begin{equation}\label{eq:hahnpoly}
	\begin{array}{c}
		k_p^{(c)}\left( {s,a,b} \right) = \frac{{{{(a - b + 1)}_p}{{(a + c + 1)}_p}}}{{p!}} \times {}_3{F_2}( - p,a - s,a + s + 1;a - b + 1,a + c + 1;1)\\
		{\rm{for}}\;p,x = 0,1, \ldots ,N - 1,s = a,a + 1, \ldots ,b - 1
	\end{array}
\end{equation}

The polynomials are on the interval $(a,b-1)$ and calculated using the additional parameter $c$.  $a$, $b$, and $c$ are subject to the constraints
\[
- \frac{1}{2} < a < b,\quad \left| c \right| < 1 + a,\quad b = a + N
\]
where $a$ determines the origin of the lattice, and the difference between $a$ and $c$ affects the range of polynomial values. The larger the difference between $a$ and $c$, the larger the range of polynomial values will be. The weight function is given by
\begin{equation}\label{eq:hahnweight}w(s)= \frac{\Gamma{(a+s+1)}\Gamma{(c+s+1)}}{\Gamma{(s-a+1)}\Gamma{(b-s)}\Gamma{(b+s+1)}\Gamma{(s-c+1)}}\end{equation}
and the squared norm is 
\begin{equation}\label{eq:hahnnorm}\rho(p)= \frac{\Gamma{(a+c+p+1)}}{p!(b-a-p-1)!\Gamma{(b-c-p)}}\end{equation}
The normalized dual Hahn polynomial is defined as 
\begin{equation}\label{eq:hahnnormpoly}\hat{k}_p^{(c)}(s,a,b)= k_p^{(c)}(s,a,b)\sqrt{\frac{w(s)}{\rho(p)} (2s+1)}\end{equation}
Dual Hahn polynomials follow the recurrence formula
\begin{equation}\label{eq:hahnrecur}\hat{k}_{p+1}^{(c)}(s,a,b)= A(\frac{\rho(p)}{\rho(p+1)})\hat{k}_{p}^{(c)}(s,a,b)- B(\frac{\rho(p-1)}{\rho(p+1)})\hat{k}_{p-1}^{(c)}(s,a,b)\end{equation}
where 
\[
\begin{split}
	A&=\frac{1}{p+1}(s(s+1)- ab + ac-bc-(b-a-c-1)(2p+1)+2p^2 ),\\
	B&=-\frac{1}{p+1}(a+c+p)(b-a-p)(b-c-p)
\end{split}
\]
and 
\[
\begin{split}
	\hat{k}_0^{(c)}(s,a,b)&= \sqrt{\frac{w(s)}{\rho(0)}(2s+1)},\\
	\hat{k}_1^{(c)}(s,a,b)&=((s-a)(s+b)(s-c)-(a+s+1)(c+s+1)(b-s-1))\sqrt{\frac{w(s)}{\rho(1) (2s+1)}}
\end{split}
\]

\subsection{Polar transform}
Moment invariants derived from rectangular orthogonal polynomials with respect to translation are straightforward to calculate. Still, invariants with respect to rotation are difficult to derive \cite{Flusser2009}. The \pkg{IM} package deals with the problem of rotation invariance for orthogonal moments by transforming the image before calculating moments. The method \code{polarTransform} available in the \pkg{IM} package essentially "unwraps" the image from the centroid or center of the image frame. More specifically, the image is represented by an intensity map based on the polar coordinates of the pixels. The vertical axis is the pixel radius from the centroid $r$, the horizontal axis is the pixel angle $theta$, and the value at location $(r,theta)$ in the heat map is the pixel intensity. The \code{polarTransform} method treats the image as a series of concentric circles of pixels around the centroid and divides each ring into equally spaced intervals. A parameter, $resolution$, must be supplied to the method when it is called. The resolution determines how many intervals between $0$ and $2\pi$ will be represented in the transformed image. In order to construct this transform, the range of radius and angle values are calculated first. The radius is ranged from 0 to the radius of the largest size circle, which may be confined within the image boundaries. The angle is ranged from 0 to $2\pi$, discretized into evenly spaced intervals. Each circle around the image's centroid will be discretized into $r*resolution$ intervals between 0 and $2\pi$, where $r$ is the radius of each circle. The location of angle 0 will be shifted to coincide with the principal axis of the image. The principal axis $\theta_0$ is determined by \cite{Flusser2009}
\begin{equation} \label{eq:praxis}\theta_0 = \frac{1}{2}\arctan\left(\frac{2\mu_{00}}{\mu_{10}-\mu_{01}}\right)\end{equation}
where 
\begin{equation} \label{eq:imagecenter}\mu_{pq} = \sum_{1}^{N} \sum_{1}^{M}(x-\bar{x})^p(y-\bar{y})^q f(x,y)\end{equation}
$(\bar{x},\bar{y})$ are the coordinates of the centroid. Next, the pixel's intensity, whose polar coordinates most closely match the coordinates of each point in the heat map, is used as the point value. If the parameter \code{center} is supplied, specifying the coordinates of the image centroid, the transform will be performed about those coordinates instead of the center of the image frame. 

After this transform is completed, polynomials that are orthogonal over a rectangle may be used to calculate moments invariant to rotation. The function \code{revPolar} is also provided. It performs an inverse transform to recover the original image transformed by the method \code{polarTransform}. A larger \code{resolution} used when transforming the image will result in a better reconstruction by the \code{revPolar} method.

\begin{figure}[H]
	\centering
	\begin{subfigure}[t]{0.30\textwidth}
		\centering
		\includegraphics[width=\textwidth]{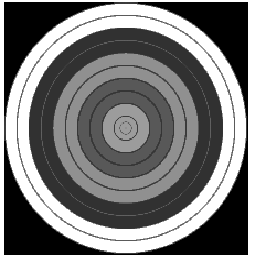}
	\end{subfigure}
	~
	\begin{subfigure}[t]{0.30\textwidth}
		\centering
		\includegraphics[width=\textwidth]{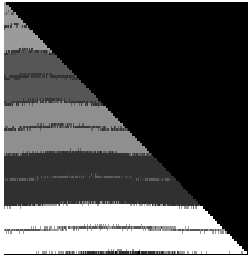}
	\end{subfigure}
	~
	\begin{subfigure}[t]{0.30\textwidth}
		\centering
		\includegraphics[width=\textwidth]{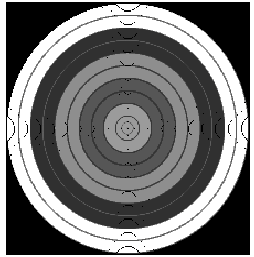}
	\end{subfigure}
	\caption{Illustration of polar transformation using the \pkg{IM} package. The original \code{circles} image (left), the transformed image (center), and the reverse-transformed image (right).}
	\label{fig:polar}
\end{figure}

The image \code{circles}, available as sample data in the \pkg{IM} package, represents a series of concentric circles and is used in the following example to illustrate the use of the \code{polarTransform} method (Fig. \ref{fig:polar}). The following code example is found in the \code{demo_polarTransform} demo provided with the package.
\begin{CodeChunk}
	\begin{CodeInput}
		R> data("circles")
		R> img <- rowSums(img, dim = 2)
		R> transf <- polarTransform(img, resolution = 20, scale = 100, 
		+    center = calcCentroid(img))
		R> itransf <- revPolar(dim(img), transf);
		R> displayImg(list(img, transf[[1]], itransf))
	\end{CodeInput}
\end{CodeChunk}

\section{Continuous complex moments}
Complex moments are calculated in the polar coordinate space of the image using complex polynomials. Because of this transformation,
acquiring moment invariants with respect to rotation is done by taking the magnitude of the moments \cite{Flusser2009, Mukundan1998, Ping2002}. To calculate complex moments, since they are orthogonal over the unit disk, the euclidean coordinates of pixels in the image need to be transformed to polar coordinates and scaled to the unit disk \cite{Bayraktar2006, Flusser2009, Mukundan1995, Mukundan1998, Mukundan2004, Ping2002, Sheng1994}. This coordinate transform is implemented in the \pkg{IM} package as the function \code{polarXY}. The general equations for calculating moments and reconstructing images are similar to those used for polynomials orthogonal over a rectangle. Complex moments $M_{p\lambda}$ of an image are calculated using the following equation:
\begin{equation}
	\begin{split}
		\label{cmplxInt}M_{p\lambda} \int^{2\pi}_0 \int^1_0 k_{p\lambda}(r)e^{-i\lambda\theta}f(r,\theta)r\,dr\,d\theta\\
		\text{for } p=0,1,2,\dots \text{ and } \lambda = -p,\dots,p
	\end{split}
\end{equation}

The polar coordinates of the pixels in the image are represented by radius $r$ and angle $\theta$. The order of the polynomials is represented by the parameter $p$. A second variable, the repetition parameter $\lambda$, is used in addition to the order for calculating complex polynomials. 

The complex polynomial is represented by $k_{p\lambda}$, the radial part of the polynomial, and $e^{-i\lambda\theta}$, the angular part of the polynomial. Like the orthogonal polynomials described previously, the polynomials are a function of the pixel coordinates, except that they are a function of the pixel polar coordinates instead of their euclidean coordinates. The \pkg{IM} package approximates the integral in Equation \ref{cmplxInt} using the summation:
\begin{equation}\label{eq:complexmomentapprox}M_{p\lambda} \sum_{x=1}^{N} \sum_{y=1}^{M} k_{p\lambda}(r_{xy})e^{-i\lambda\theta_{xy}}f(x,y)\end{equation}
where $r_{xy}$ and $\theta_{xy}$ are the radius and angle of the pixel located at $(x,y)$.\\
Complex moments are symmetric with respect to repetition \cite{Flusser2009,Mukundan1995,Mukundan1998}, so only half of the moments need to be calculated;  the other half are simply equal to the complex conjugate of the calculated moments. They are related by the equation
\begin{equation}\label{eq:conjmoments}M_{p,\lambda} = \textrm{conj}(M_{p,-\lambda}).\end{equation}
Reconstruction of the original image from the moments is defined by the following equation \cite{Flusser2009,Mukundan1998}:
\begin{equation}\label{eq:complexreconstruct}\hat{f}(x,y) = \sum_{p=0}^{\infty}\sum_{\lambda=-p,-p+2,…}^{p}k_{p\lambda}(r_{xy})e^{-i\lambda\theta_{xy}}M_{p\lambda}\end{equation}
The maximum order used to reconstruct the image can be arbitrarily large.

\subsection{Generalized pseudo-Zernike polynomials} 

\begin{figure}[H]
	\centering
	\includegraphics[width=\textwidth]{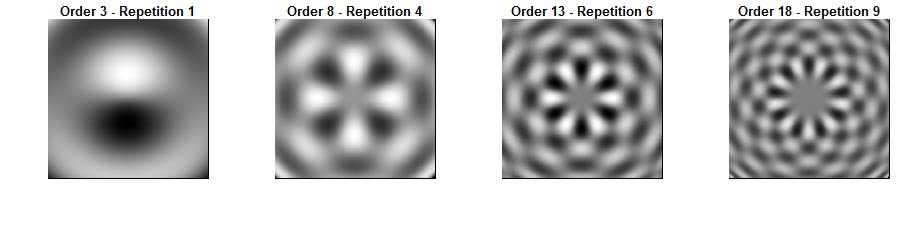}
	\caption{Generalized pseudo-Zernike kernels with $\alpha=0$ and varying order and repetition.}
	\label{fig:zernike}
\end{figure}

Generalized pseudo-Zernike polynomials (Fig. \ref{fig:zernike}) $k_{p\lambda}$ of order $p$ and repetition $\lambda$ are formally defined as \cite{Bayraktar2006,Flusser2009,Mukundan1995,Mukundan1998}
\begin{equation}\label{eq:gpzmpoly}k_{p\lambda}^{\alpha}(z)= z^{\frac{p+\lambda}{2}}(z^{*})^{\frac{p-\lambda}{2}} \frac{(\alpha+1)_{p-|\lambda|}}{(p-|\lambda|)!}{}_2 F_1\left(-p+|\lambda|, -p -|\lambda|-1;\alpha+1; 1- \frac{1}{(zz^*)^{1/2}}\right)\end{equation} 
where $*$ denotes the complex conjugate and
\[z= r \text{ exp}^{j\theta}\]
$\alpha$ is a user-defined parameter that scales the polynomial values. A larger value of $\alpha$ will decrease the range of the polynomial values. The repetition $\lambda$ is constrained by order \emph{p}:
\begin{equation}
	\label{repConstraint}0 \le |\lambda| \le p.
\end{equation}

In terms of polar coordinates, the polynomial is defined as 
\begin{equation}\label{eq:gpzmpolypolar}k_{p\lambda}^{\alpha}(r,\theta)= k_{p\lambda}^{\alpha}(r \text{ exp}(j\theta))= R_{p\lambda}^{\theta}(r)\text{exp}(j\lambda\theta),\end{equation}
where the real-valued radial polynomial $R_{p\lambda}^{\alpha}(r)$ is given by
\begin{equation}\label{eq:gpzmpolyradial}R_{p\lambda}^{\alpha}(r)= \frac{(p+|\lambda|+1)!}{(\alpha+1)_{p+|\lambda|+1}}\sum_{s=0}^{p-|\lambda|} \frac{(-1)^s(\alpha+1)_{2p+1-s}}{s!(p-|\lambda|-s)!(p+|\lambda|+1-s)!}r^{p-s}\end{equation}
When $a=0$, generalized pseudo-Zernike polynomials are reduced to pseudo-Zernike polynomials.
The weight function is 
\begin{equation}\label{eq:gpzmweight}w(r)=(1-r)^{\alpha}\end{equation}
and the norm is 
\begin{equation}\label{eq:gpzmnorm}\rho_{p\lambda}^{\alpha}= \frac{2\pi (p-|\lambda|+1)_{2|\lambda|+1}}{(2p+\alpha+2)(\alpha+1+p-|\lambda|)_{2|\lambda|+1}}\end{equation}
The normalized generalized pseudo-Zernike polynomial is defined as 
\begin{equation}\label{eq:gpzmnormpoly}\hat{R}_{p\lambda}^{\alpha}(r)= R_{p\lambda}^{\alpha}\sqrt{\frac{w(r)}{\rho_{p\lambda}^{\alpha}}}.\end{equation}
The radial polynomial $R_{p\lambda}^{\alpha}(r)$ is computed in the \pkg{IM} package using the recurrence relation
\begin{equation}\label{eq:gpzmrecur}R_{p\lambda}^{\alpha}(r)= (M_1 r +M2)R_{p-1,\lambda}^{\alpha}(r)+ M_3 R_{p-2,\lambda}^{\alpha}(r),\end{equation}
with

\begin{eqnarray*}
	M_1 &=& \frac{(2p+1+\alpha)(2p+\alpha)}{(p+\lambda+1+\alpha)(p-\lambda)}\\
	M_2 &=& -\frac{(p+\lambda+1)(\alpha+2p)}{p+\lambda+\alpha+1}+ M_1\frac{(p+\lambda)(p-\lambda-1)}{2p-1+\alpha}\\
	M_3 &=& \frac{(p+\lambda)(p+\lambda+1)(2p-2+\alpha)(2p-1+\alpha)}{2(p+\lambda+\alpha+1)(p+\lambda+\alpha)}\\
	&& +M_2\frac{(p+\lambda)(2p-2+\alpha)}{p+\lambda+\alpha}-M_1\frac{(p+\lambda)(p+\lambda-1)(p-\lambda-2)}{2(p+\lambda+\alpha)}
\end{eqnarray*}

and
\begin{eqnarray*}
	R_{\lambda\lambda}^{\alpha}(r) &=& r^\lambda\\
	R_{\lambda+1,\lambda}^{\alpha}(r) &=& [(\alpha+3+2\lambda)r-2(\lambda+1)]R_{\lambda\lambda}^{\alpha}(r)
\end{eqnarray*}

\subsection{Fourier-Mellin polynomials} 
\begin{figure}[H]
	\centering
	\includegraphics[width=\textwidth]{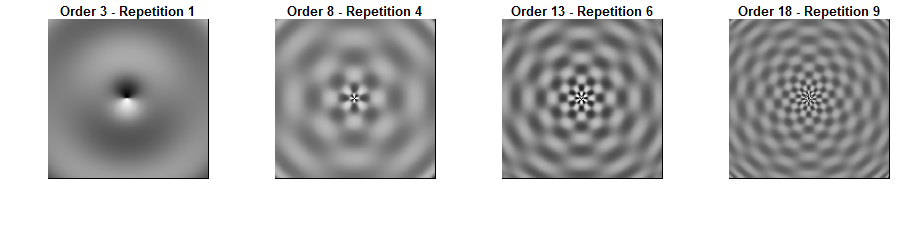}
	\caption{Fourier-Mellin kernels with varying order and repetition.}
	\label{fig:fourier-mellin}
\end{figure}

Fourier-Mellin polynomials (Fig. \ref{fig:fourier-mellin}) of order $p$ and repetition $\lambda$ are defined as \cite{Sheng1994}
\begin{equation}\label{eq:fmpoly}V_{p\lambda}(r,\theta)= R_p(r)\text{ exp}(j\lambda\theta)\end{equation} 
with the radial polynomial $R_p(r)$ expressed as
\begin{equation}\label{eq:fmradialpoly}R_p{r}= \sum_{k=0}^{p} (-1)^{n+k}\frac{(n+k+1)!}{k!(k+1)!(n-k)!}r^k\end{equation}
When the real-valued radial part of a generalized pseudo-Zernike polynomial has parameters $\alpha=0$ and $\lambda=0$, it is reduced to the Fourier-Mellin polynomial $R_p(r)$. The Fourier-Mellin polynomial is computed recursively by
\begin{equation}\label{eq:fmrecur}R_p(r)= (M_1 r+ M_2)R_{p-1}(r) + M_3 R_{p-2}(r)\end{equation}
with
\begin{eqnarray*}
	M_1 &=& \frac{2(2p+1)}{(p+1)}\\
	M_2 &=& -2p+ M_1\frac{p(p-1)}{2p-1}\\
	M_3 &=& (p-1)(2p-1)+ M_2(2p-2)-M_1\frac{(p-1)(p-2)}{2}
\end{eqnarray*}
and
\begin{eqnarray*}
	R_{0}^{\alpha}(r) &=& 1\\
	R_{1}^{\alpha}(r) &=& 3r-2
\end{eqnarray*}

\subsection{Chebyshev-Fourier polynomials} 
\begin{figure}[H]
	\centering
	\includegraphics[width=\textwidth]{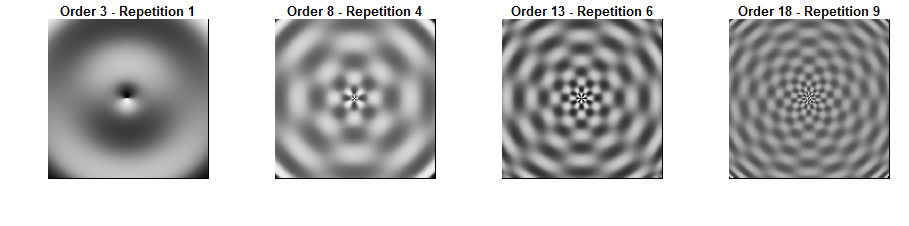}
	\caption{Chebyshev-Fourier kernels with varying order and repetition.}
	\label{fig:chebyshev-fourier}
\end{figure}

Chebyshev-Fourier polynomials (Fig. \ref{fig:chebyshev-fourier}) of order $p$ and repetition $\lambda$ are defined by \cite{Ping2002} as
\begin{equation}\label{eq:cfpoly}k_{p\lambda}(r,\theta)= R_p(r)\text{ exp}(j\lambda\theta)\end{equation} 
The radial polynomial $R_p(r)$ is a normalized and shifted Chebyshev polynomial of the second kind \cite{Flusser2009} weighted by the function
\begin{equation}\label{eq:cfweight}w(r)= (r-r^2)^{1/2}\end{equation}
Hence, it can be expressed explicitly as
\begin{equation}\label{eq:cfnormpoly}R_p(r)= \sqrt{\frac{8}{\pi}}\left(\frac{1-r}{r}\right)^{1/4}\sum_{\frac{p+2}{2}}^{k=0} (-1)^k\frac{(p-k)!}{k!(p-2k)!}[2(2r-1)]^{p-2k}\end{equation}

\subsection{Radial Harmonic Fourier function}
\begin{figure}[H]
	\centering
	\includegraphics[width=\textwidth]{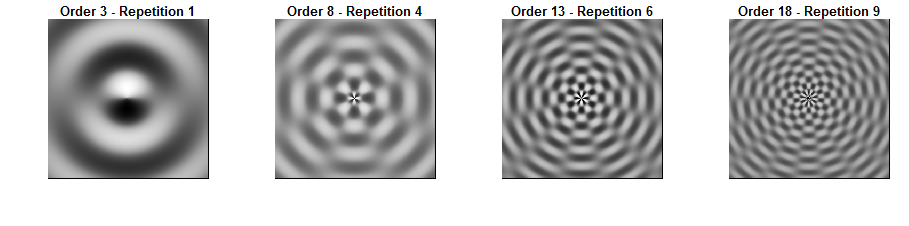}
	\caption{Radial harmonic Fourier kernels with varying order and repetition.}
	\label{fig:harmonic}
\end{figure}

The radial harmonic Fourier function (Fig. \ref{fig:harmonic}) is not a polynomial in $r$, but it is orthogonal on $[0,1]$ \cite{Flusser2009}. It is defined as
\begin{equation}\label{eq:rhfpoly}k_p(r)=\left\{\begin{matrix}
		\frac{1}{\sqrt{r}} &  \text{if }p=0,
		\\ 
		\sqrt{\frac{2}{r}}\sin[(p+1)\pi r] & \text{if } p=\text{odd},
		\\ 
		\sqrt{\frac{2}{r}}\cos(p\pi r) & \text{if } p=\text{even}.
	\end{matrix}\right.
\end{equation}

\section[The IM package]{The \pkg{IM} package}
The \pkg{IM} package includes methods that calculate several types of orthogonal moments and moment invariants of images and perform reconstruction of images from moments. It contains three classes: \code{CmplxIm}, \code{OrthIm}, and \code{MultiIm}. The \code{CmplxIm} class may be used for calculating continuous complex moments of an image. The \code{OrthIm} class may be used for calculating discrete or continuous orthogonal moments of an image. Methods to transform an image by ``polar unwrapping'' before calculating orthogonal moments are available. The \code{MultiIm} class may be used for calculating any of the available types of moments for multiple images with the same dimensionality. This class performs faster by re-using a single set of polynomials for all images. Moments and polynomials may also be calculated without using the available classes. However, the use of the classes is recommended to ensure that all necessary parameters are set correctly.

\subsection[The OrthIm class]{The \code{OrthIm} class}
The \code{OrthIm} class available in the \pkg{IM} package is intended for use in calculating real discrete and continuous moments whose polynomials are orthogonal on a rectangle or square. The continuous orthogonal moments available in this package are \emph{continuous Chebyshev}, \emph{Legendre}, and \emph{Gegenbauer} moments. The discrete orthogonal moments available are \emph{discrete Chebyshev}, \emph{dual Hahn}, and \emph{Krawtchouk} moments. An \code{OrthIm} object may be initialized using the standard \proglang{R} constructor \code{new()} \cite{CRAN}.

\begin{CodeChunk}
	\begin{CodeInput}
		R> obj <- new("OrthIm", filename = f, img = i)
	\end{CodeInput}
\end{CodeChunk}

The object may be initialized in one of three ways. If no arguments are provided for \code{filename} and \code{img}, an S4 object will be created containing the empty variable slots of an \code{OrthIm} object. If a filename is provided, the constructor will import the image, convert it to grayscale, and return an object with slots \code{I, dimensions, centroid, filename}, and \code{fileType} initialized. This option is available only for jpeg, bmp, and png image files. The packages \pkg{png} (>= 0.1-4) \cite{png}, \pkg{jpeg} (>= 0.1-2) \cite{jpeg}, and \pkg{bmp} (>= 0.1) \cite{bmp} are required in order for jpeg, bmp, and png image files to be loaded into \pkg{R}. Alternatively, an image matrix may be provided as the input argument \code{i} in the above example, which will result in a new \code{OrthIm} object with slots \code{I, dimensions}, and \code{centroid} initialized. Slot \code{I} in the object is the image matrix for which moments are to be calculated.

This class has methods that may be used to set necessary parameters and calculate moments of any of the types mentioned above, as well as combinations of any two moments. If two types of moments are combined, both must be either continuous or discrete \cite{Zhu2012}. The type of moments to be calculated can be defined in the following way:

\begin{CodeChunk}
	\begin{CodeInput}
		R> momentType(obj) <- "cheby"
	\end{CodeInput}
\end{CodeChunk}

Where \code{cheby} represents \emph{discrete Chebyshev}. 
If two types of polynomials are to be used, such as a combination of \emph{Legendre} polynomials in the horizontal direction and \emph{Gegenbauer} polynomials in the vertical direction, the following code may be used:

\begin{CodeChunk}
	\begin{CodeInput}
		R> momentType(obj) <- c("legend", "gegen")
	\end{CodeInput}
\end{CodeChunk}

The maximum order of moments to be calculated in the vertical and horizontal directions must be set in the following way:

\begin{CodeChunk}
	\begin{CodeInput}
		R> setOrder(obj) <- c(orderX, orderY)
	\end{CodeInput}
\end{CodeChunk}

A single value may be supplied as the input argument if the maximum orders are the same in both directions. Additional parameters may be required depending on the type of moments to be calculated. These would be specified using the \code{setParams} method. 

An \code{OrthIm} object will store the polynomials used for calculating the moments. The polynomials may be re-used to calculate moments for other images with the same dimensionality. This technique is used in the \code{MultiIm} class for processing multiple images. A method to perform image reconstruction from any maximum order of moments less than or equal to those calculated is also available in both \code{OrthIm} and \code{MultiIm} classes.

To demonstrate the \code{OrthIm} class, bivariate moments of the sample data image \code{mandril} will be calculated. Continuous Chebyshev polynomials of order 100 in the $x$ direction and Legendre polynomials of order 150 in the $y$ direction will be used to calculate the bivariate moments. The following code is available as the demo \code{demo_OrthIm}.

\begin{CodeChunk}
	\begin{CodeInput}
		R> data("mandril")
		R> obj <- new("OrthIm", img = img, filename = "")
		R> momentType(obj) <- c("chebycont", "legend")
		R> setOrder(obj) <- c(150, 150)
		R> Moments(obj) <- NULL
	\end{CodeInput}
\end{CodeChunk} 

Instead of using an \code{OrthIm} object, the function \code{momentObj} may be used to accomplish the same task as the example above in the following way:

\begin{CodeChunk}
	\begin{CodeInput}
		R> data("mandril")
		R> obj <- momentObj(img, c("chebycont", "legend"), c(150, 150), NULL, c())
	\end{CodeInput}
\end{CodeChunk}

\begin{figure}[H]
	\centering
	\begin{subfigure}[t]{0.30\textwidth}
		\centering
		\includegraphics[width=\textwidth]{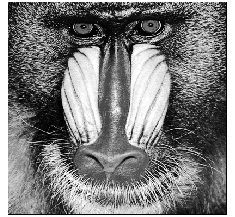}
	\end{subfigure}
	~
	\begin{subfigure}[t]{0.30\textwidth}
		\centering
		\includegraphics[width=\textwidth]{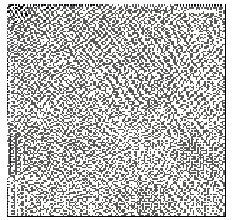}
	\end{subfigure}
	~
	\begin{subfigure}[t]{0.30\textwidth}
		\centering
		\includegraphics[width=\textwidth]{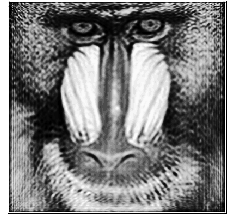}
	\end{subfigure}
	\caption{The images displayed in the above code example using the function \code{displayImg}. The original \code{mandril} image (left), continuous Chebyshev-Legendre moments of the image up to orders $(150,150)$ (center) and the reconstruction of the \code{mandril} image using Chebyshev-Legendre moments (right).}
	\label{fig:mandril}
\end{figure}

Reconstruction is performed using the \code{Reconstruct} method in the following example. The original image, moment matrix, and reconstructed image are displayed (Fig. \ref{fig:mandril}).

\begin{CodeChunk}
	\begin{CodeInput}
		R> Reconstruct(obj) <- c(125, 125)
		R> displayImg(list(obj@I, obj@moments, obj@reconstruction))
	\end{CodeInput}
\end{CodeChunk}

\subsection[The CmplxIm class]{The \code{CmplxIm} class}
The \code{CmplxIm} class available in the \pkg{IM} package is intended for use in calculating continuous complex moments that are orthogonal on the unit disk. The continuous complex moments available in this package are \emph{generalized pseudo-Zernike}, \emph{Fourier-Mellin}, \emph{Chebyshev-Fourier}, and \emph{radial harmonic Fourier} moments. A \code{CmplxIm} object may be initialized using the standard \proglang{R} \cite{CRAN} constructor \code{new()}.

\begin{CodeChunk}
	\begin{CodeInput}
		R> obj <- new("CmplxIm", filename = f, img = i)
	\end{CodeInput}
\end{CodeChunk}

The object may be initialized in the same way as an object of the \code{OrthIm} class. There are minor differences in the constructor implementation since complex moment calculation requires a scaled polar coordinate transform of the pixels. The polar coordinates of the pixels in the image are calculated when the object's image slot \code{I} is initialized. This class has methods that may be used to set necessary parameters and calculate moments of any of the types mentioned above. Unlike the orthogonal moments described in the previous section, combinations of different types of complex polynomials are not available. The type of moments to be calculated can be defined in the same way as for the \code{OrthIm} class, using the \code{momentType} method.

\begin{CodeChunk}
	\begin{CodeInput}
		R> momentType(obj) <- "gpzm"
	\end{CodeInput}
\end{CodeChunk}

\code{"gpzm"} represents generalized pseudo-Zernike. The maximum order of moments to be calculated must be set in the following way:

\begin{CodeChunk}
	\begin{CodeInput}
		R> setOrder(obj) <- order
	\end{CodeInput}
\end{CodeChunk}

Additional parameters may be required depending on the type of moments to be calculated. These can be specified using the \code{setParams} method. The method \code{Invariant} obtains moment invariants in one step by calculating the magnitude of the complex moments. If moments have not already been computed prior to using this method, they will be calculated first. The moment invariants are stored in the object slot \code{invariant}.

\begin{CodeChunk}
	\begin{CodeInput}
		R> Invariant(obj) <- NULL
	\end{CodeInput}
\end{CodeChunk}

A \code{CmplxIm} object does not store the polynomials used for calculating the moments because the polynomials are a function of $p$, $\lambda$, and $r_{xy}$ (refer to the section in this paper on Continuous Complex Moments), requiring a 4-dimensional matrix (or 3-dimensional at the cost of increased code complexity). The \code{MultiIm} class similarly does not require the polynomials to be stored in order to expedite the processing of multiple images. Still, it does assume that the centroids of all images are located at the center of an image frame. A method to perform image reconstruction from any maximum order of moments less than or equal to those calculated is also available in the \code{CmplxIm} class.

The intended use of the \code{CmplxIm} class is demonstrated in the following example. Generalized pseudo-Zernike moments of the sample data image \code{earth} are computed up to order $100$ with parameter $\alpha$ set to $1$. This code is found in the demo \code{demo_CmplxIm}.

\begin{CodeChunk}
	\begin{CodeInput}
		R> data("earth")
		R> obj <- new("CmplxIm", img = img)
		R> momentType(obj) <- "gpzm"
		R> setOrder(obj) <- 100
		R> setParams(obj) <- 1
		R> Moments(obj) <- NULL
	\end{CodeInput}
\end{CodeChunk}

Instead of using a \code{CmplxIm} object, the function \code{momentObj} may be used to accomplish the same task as the example above. A \code{CmplxIm} object will be returned.

\begin{CodeChunk}
	\begin{CodeInput}
		R> data("earth")
		R> obj <- momentObj(img, "gpzm", 100, 1)
	\end{CodeInput}
\end{CodeChunk}

\begin{figure}[H]
	\centering
	\begin{subfigure}[t]{0.30\textwidth}
		\centering
		\includegraphics[width=\textwidth]{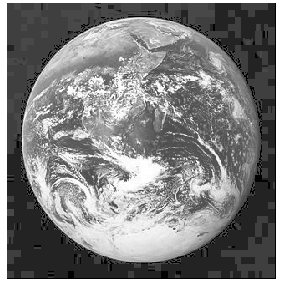}
	\end{subfigure}
	~
	\begin{subfigure}[t]{0.30\textwidth}
		\centering
		\includegraphics[width=\textwidth]{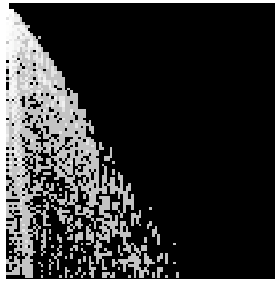}
	\end{subfigure}
	~
	\begin{subfigure}[t]{0.30\textwidth}
		\centering
		\includegraphics[width=\textwidth]{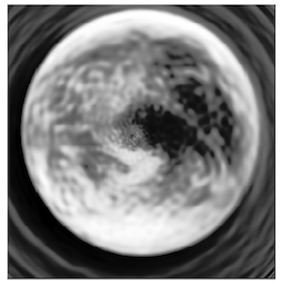}
	\end{subfigure}
	\caption{The images displayed in the above code example using the \code{displayImg} function. The original \code{earth} image (left), generalized pseudo-Zernike moments up to order $100$ of the image (center), and the reconstruction of the \code{earth} image using generalized pseudo-Zernike moments (right).}
	\label{fig:earth-reconstruction}
\end{figure}

Reconstruction may be performed by the \code{Reconstruct} method up to a specified order. In the following example, the image is reconstructed using moments up to order 80. Then, the original image, moments, and reconstructed image are displayed (Fig. \ref{fig:earth-reconstruction}).

\begin{CodeChunk}
	\begin{CodeInput}
		R> Reconstruct(obj) <- c(80, 80)
		R> displayImg(list(obj@I, abs(obj@moments), obj@reconstruction))
	\end{CodeInput}
\end{CodeChunk}

\subsection[The MultiIm class]{The \code{MultiIm} class}
The \code{MultiIm} class is used to compute moments of multiple images that have the same dimensionality, and assumes that the centroid is always at the center of the image frame. Any moment type available in the package can be computed using this class. A \code{MultiIm} object may be initialized using the standard \proglang{R} constructor \code{new()} \cite{CRAN}.

\begin{CodeChunk}
	\begin{CodeInput}
		R> obj <- new("MultiIm", images)
	\end{CodeInput}
\end{CodeChunk}

The input argument \code{images} is a list of images in 2-dimensional matrix form. The images must all be the same size. The type of moment is set using the \code{momentType} method. The order of moments to be calculated is set using the \code{setOrder} method. Any additional parameters required for the specific moment type are set using the \code{setParams} method.

\begin{CodeChunk}
	\begin{CodeInput}
		R> momentType(obj) <- c("type1", "type2")
		R> setOrder(obj) <- c(orderX, orderY)
		R> setParams(obj) <- list(paramX, NULL)
	\end{CodeInput}
\end{CodeChunk}

Moment calculation and reconstruction can be computed faster by re-using polynomial values if several images of identical size and centroid are processed. When computing real orthogonal moments (continuous or discrete), a \code{MultiIm} object calculates polynomials once, stores them, and calculates moments for each image simply by using matrix multiplication as in Equation \ref{matMoments}. When computing complex moments, a \code{MultiIm} object by default does not store polynomial values since a 4-dimensional matrix would be required to do so. However, since all the images are assumed to have the same dimensionality and centroid location, the polar-coordinate representation of the pixels may be re-used for every image. 

Moments, moment invariants, and reconstruction are computed by calling the same methods as in the \code{CmplxIm} and \code{OrthIm} classes. The calculations will be performed for all images in the image list, and the slot \code{moments} of the object will contain a list of the moments for each image. The same is true for the \code{invariant} slot of the object. The \code{momentObj} method can also handle a list of images as an input parameter and return an object of the \code{MultiIm} class. Orthogonal and complex moment types are treated the same way as in the \code{CmplxIm} and \code{OrthIm} classes. 

A \code{MultiIm} object is used in the classification examples in the next section of this document. Please see the examples there for a demonstration of the usage of the \code{MultiIm} class.

\section{Moment invariants as features for classification}

\subsection{Classification of bacterial colonies using complex moment invariants}
This example uses moment invariants of light-scatter patterns of {\it Salmonella} colonies in order to identify different bacterial serotypes. These images were collected using a laser scatterometer at Purdue University \cite{Rajwa2010}. These data are provided in the \pkg{IM} package and are entitled \code{bacteria}. Images of 4 different serotypes of {\it Salmonella} -- Agona, Hadar, Newport, and Typhimurium -- are present in the provided data set. Five examples of each of the four serotypes are given. This example is included in the \pkg{IM} package as the demo \code{cmplxClassification}. The data are loaded into the workspace in the following way:

\begin{CodeChunk}
	\begin{CodeInput}
		R> data("bacteria")
	\end{CodeInput}
\end{CodeChunk}

\begin{figure}[H]
	\centering
	\includegraphics[width=0.5\textwidth]{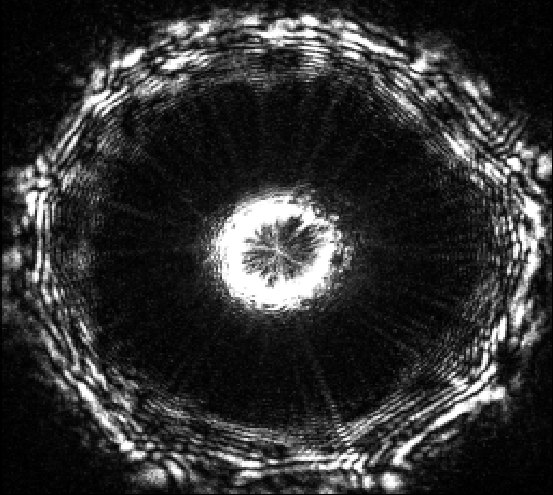}
	\caption{Laser light-scatter pattern formed by a {\it Salmonella} ser. Agona bacterial colony.}
\end{figure}

When the data are loaded, two variables appear in the workspace. The variable \code{img} is the list of images to be classified, and the variable \code{labels} is the numeric label of the class of the corresponding image in the list. First, features will be extracted from the images by calculating the moment invariants. A \code{MultiIm} object will be used to store and compute moments for all the images. Since the images are circular, complex moments will be used to achieve rotational invariance. In this example, generalized pseudo-Zernike moments will be calculated up to order 50, with the scaling parameter $\alpha$ set to 2.

\begin{CodeChunk}
	\begin{CodeInput}
		R> obj <- new("MultiIm", images = img)
		R> momentType(obj) <- "gpzm"
		R> setOrder(obj) <- 50
		R> setParams(obj) <- 2
	\end{CodeInput}
\end{CodeChunk}

To calculate moment invariants for all the images, the method \code{Invariant} will be called.

\begin{CodeChunk}
	\begin{CodeInput}
		R> Invariant(obj) <- NULL
	\end{CodeInput}
\end{CodeChunk}

The \code{invariants} slot of the object, \code{obj}, will now contain a list of matrices of moment invariants. Each matrix contains the moment invariants for one of the images. In order to use this information in a classifier, it is helpful to turn each matrix into a vector so that the input to the classifier will be a single matrix, with one row for each image and columns representing the features (moment invariants) of each image. An additional consideration is that only the lower triangular and diagonal of the invariant matrices for each image should be vectorized since the rest of the matrix will contain zeros. The reason is that generalized pseudo-Zernike moments are calculated up to order $p$ and repetition $\lambda$, where $\lambda$ is constrained by $p$ as in Equation \ref{repConstraint}. In a matrix of generalized pseudo-Zernike moments with dimensions equal to $p$, the rows would correspond to the order $p$, and the columns would correspond to $\lambda$. 

\begin{CodeChunk}
	\begin{CodeInput}
		R> invariants <- list()
		R> for (i in 1:length(labels)) {
			R> invariants[[i]] <- c(obj@invariant[[i]][lower.tri(obj@invariant[[i]], 
			+    diag = TRUE)])
			R> }
		R> invariants <- t(do.call(cbind, invariants))
	\end{CodeInput}
\end{CodeChunk}

The next step of the classification process will be testing. The data must be split into testing and training groups to use a supervised classifier. In order to test the effectiveness of moment invariants as features for classification fairly, stratified random sampling is used for this process. One-third of the images from each label group will be randomly selected for training, and the rest will be used to test the model produced by the training data. 

\begin{CodeChunk}
	\begin{CodeInput}
		R> trainIndex <- logical(length(labels))
		R> uniqueY <- unique(labels)
		R> for (i in 1:length(uniqueY)) {
			R> trainIndex[sample(which(labels == uniqueY[i]), 3)] <- TRUE
			R> }
		R> trX <- invariants[trainIndex, ]
		R> trY <- labels[trainIndex]
		R> tstX <- invariants[!trainIndex, ]
		R> tstY <- labels[!trainIndex]
	\end{CodeInput}
\end{CodeChunk}

Using the moment invariants of the images selected as training data, a support vector machine (SVM) model will be trained. The SVM classifier from the package \pkg{e1071} \cite{e1071} is used in this example. 

\begin{CodeChunk}
	\begin{CodeInput}
		R> library("e1071")
		R> model <- svm(x = trX, y = as.factor(trY))
	\end{CodeInput}
\end{CodeChunk}

Finally the test data will be classified using the SVM model (Table. \ref{table:salmonella}), and misclassification error will be calculated as the proportion of images incorrectly classified.

\begin{CodeChunk}
	\begin{CodeInput}
		R> pred <- predict(model,tstX)
		R> err <- sum(tstY != pred) / length(pred)
	\end{CodeInput}
\end{CodeChunk}

\begin{table}\centering
	\begin{tabular}{ll}
		\hline \hline
		True labels & Predicted labels \\ \hline
		Agona       & Agona            \\
		Agona       & Agona            \\
		Hadar       & Hadar            \\
		Hadar       & Hadar            \\
		Newport     & Newport          \\
		Newport     & Newport          \\
		Typhimurium & Newport          \\
		Typhimurium & Typhimurium      \\
		\hline \hline
	\end{tabular}
	\caption {True labels of the test data versus the predicted labels given by the SVM classifier.}
	\label{table:salmonella}
\end{table}

The same process illustrated by the example above can also be implemented using single \code{CmplxIm} objects for each image. In that case, the moment invariants will be calculated and vectorized within a loop. The demo \code{cmplxClassificationLoop} implements this method.

\subsection{Tamil character recognition using orthogonal moment invariants}

This example performs the classification of hand-written \emph{Tamil} characters provided by HP Labs India \cite{TamilDataset}. Eight different types of characters are represented in the data set. There are 40 images of each type. Classification of these images will be based on discrete Chebyshev moments. This example is included in the \pkg{IM} package as the demo \code{orthClassification}.

\begin{figure}[H]
	\centering
	\includegraphics[width=0.95\textwidth]{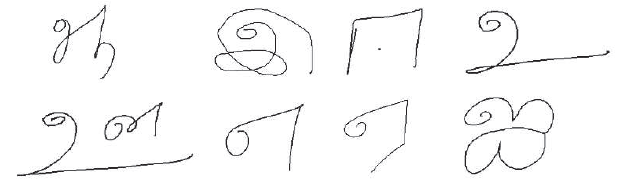}
	\caption{Examples of eight types of Tamil characters provided in the \code{characters} data set.}
\end{figure}

The first step is to load the data set \code{characters} into the workspace. When the data are loaded, two variables will appear in the workspace. The variable \code{img} is the list of images to be classified, and \code{labels} is the numeric label of the class of the corresponding image in the list. Then, initialize a new \code{MultiIm} object. Discrete Chebyshev moments of the transformed images will be calculated up to order 5 in the x and y dimensions.

\begin{CodeChunk}
	\begin{CodeInput}
		R> data("characters")
		R> obj <- new("MultiIm", images = img)
		R> momentType(obj) <- "cheby"
		R> setOrder(obj) <- c(5, 5)
		R> Moments(obj) <- NULL
	\end{CodeInput}
\end{CodeChunk}

As in the previous example, the moment invariants must be vectorized so they can be used as input to a classifier. 

\begin{CodeChunk}
	\begin{CodeInput}
		R> moments <- list()
		R> for (i in 1:length(labels)) {
			R> moments[[i]] <- c(obj@moments[[i]])
			R> }
		R> moments <- t(do.call(cbind, moments))
	\end{CodeInput}
\end{CodeChunk}

For testing, the data will be split in a random stratified manner into testing and training groups. 

\begin{CodeChunk}
	\begin{CodeInput}
		R> trainIndex <- logical(length(labels))
		R> uniqueY <- unique(labels)
		R> for (i in 1:length(uniqueY)) {
			R> trainIndex[sample(which(labels == uniqueY[i]), 26)] <- TRUE
			R> }
		R> trX <- moments[trainIndex, ]
		R> trY <- labels[trainIndex]
		R> tstX <- moments[!trainIndex, ]
		R> tstY <- labels[!trainIndex]
	\end{CodeInput}
\end{CodeChunk}

Finally an SVM model will be trained and used to predict the labels of the test data, and misclassification error will be calculated.

\begin{CodeChunk}
	\begin{CodeInput}
		R> library("e1071")
		R> model <- svm(x = trX, y = as.factor(trY))
		R> pred <- predict(model, tstX)
		R> err <- sum(tstY != pred) / length(pred)
	\end{CodeInput}
\end{CodeChunk}

The same process illustrated by the example above can also be implemented using single \code{OrthIm} objects for each image or using the \code{momentObj} function. In that case, the moment invariants will be calculated and vectorized within a loop.

\section{Reconstruction performance}
Reconstruction error for all moment types is computed using peak signal-to-noise ratio (PSNR), based on the mean-squared-error MSE.

\begin{eqnarray}\label{eq:err}
	\text{PSNR} &=& 10 \times \log_{10}\left(\frac{255^2}{MSE}\right)\\
	\text{MSE} &=& \sqrt{\frac{\sum_{x=1}^N \sum_{y=1}^M (f(x,y) - \hat{f}(x,y))^2}{N \times M}}
\end{eqnarray}

\begin{figure}
	\centering
	\includegraphics[width=0.40\textwidth]{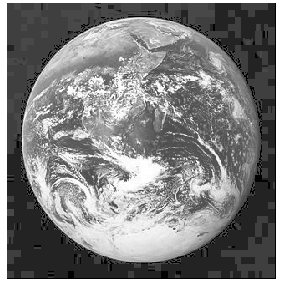}
	\caption{The \code{earth} image provided in the \pkg{IM} package.}
	\label{fig:earth}
\end{figure}

In Equation \ref{eq:err}, $f(x,y)$ is the original image and $\hat{f}(x,y)$ is the reconstructed image. Before calculating the reconstruction error, both the original and reconstructed images were normalized so that the pixel intensity values ranged between 0 and 1. The sample image used for reconstruction comparison is \code{earth}, shown in figure \ref{fig:earth}. 

\begin{figure}
	\centering
	\includegraphics[width=\textwidth]{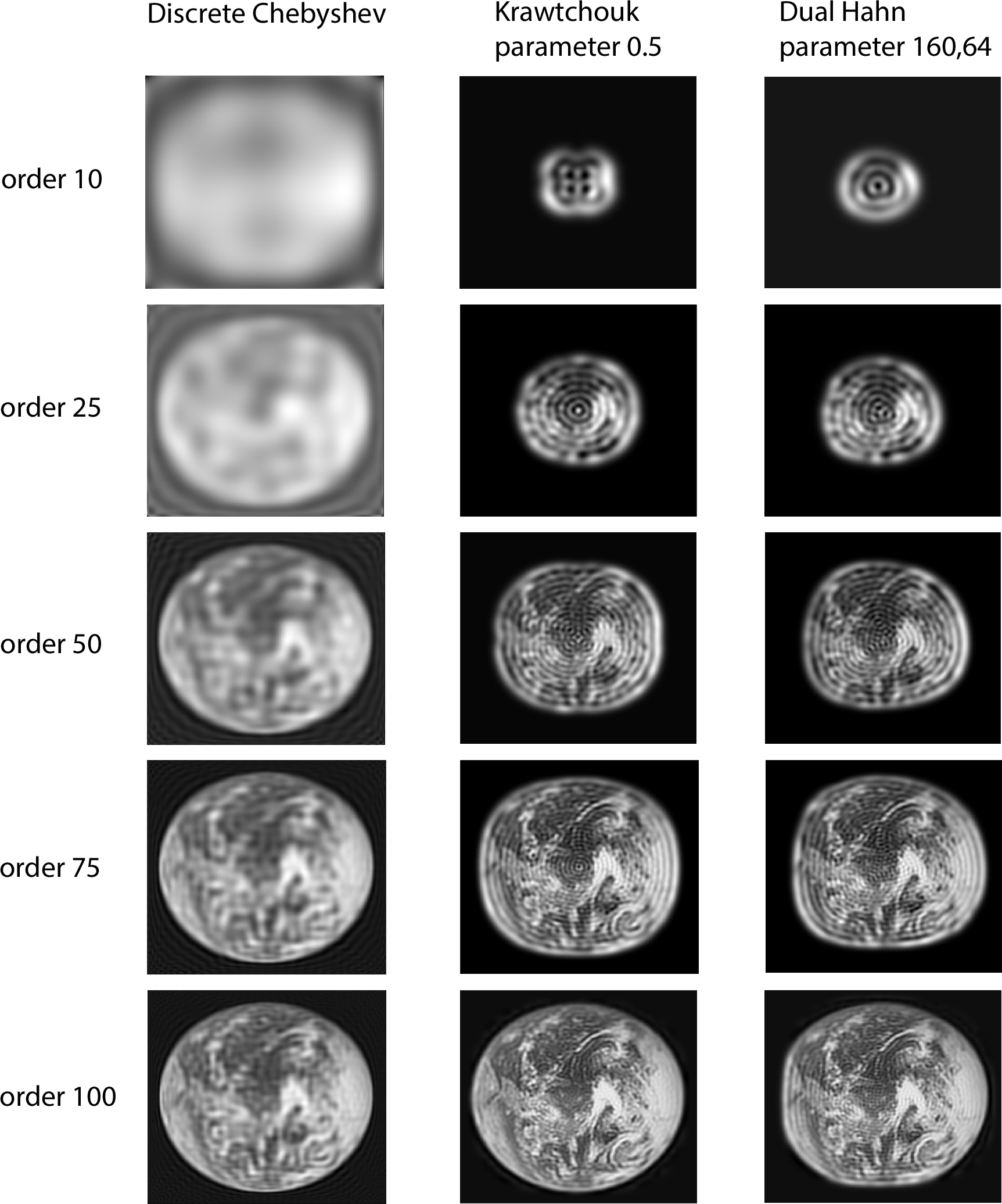}
	\caption{Reconstructions of the \code{earth} image using discrete orthogonal moments.}
	\label{fig:orthDiscReconstruct}
\end{figure}

\begin{figure}
	\centering
	\includegraphics[width=\textwidth]{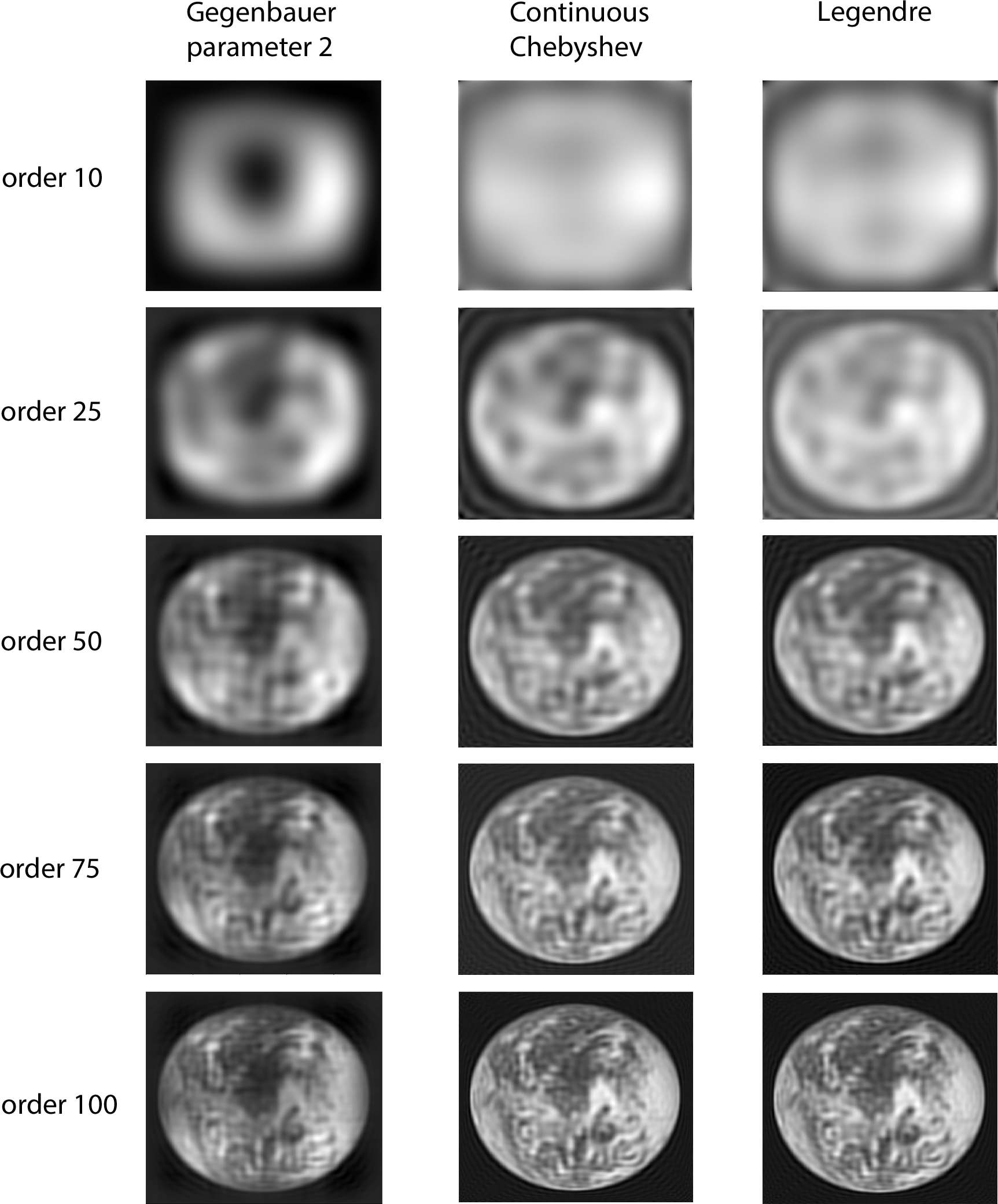}
	\caption{Reconstructions of the \code{earth} image using continuous orthogonal moments.}
	\label{fig:orthContReconstruct}
\end{figure}

\begin{figure}
	\centering
	\includegraphics[width=\textwidth]{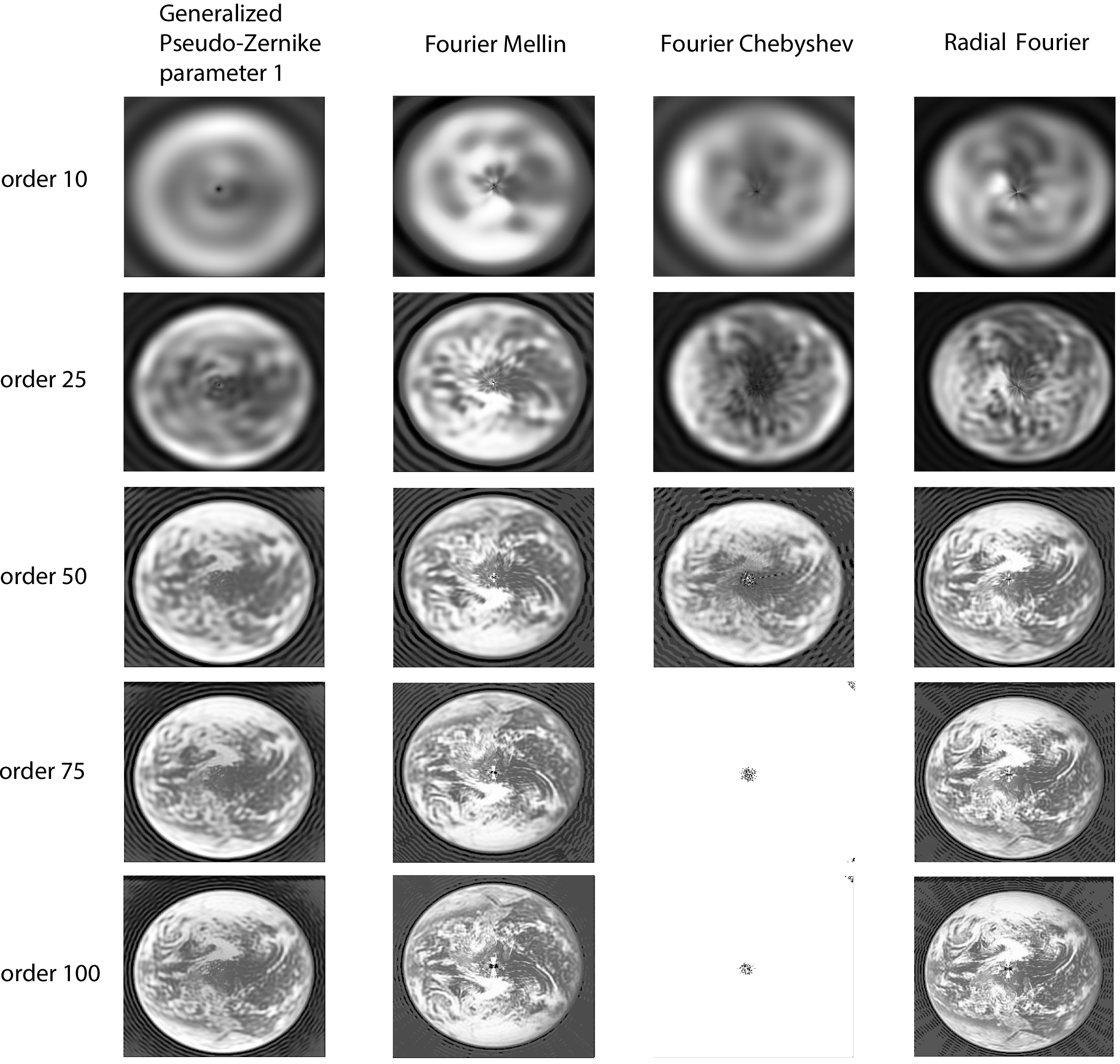}
	\caption{Reconstructions of the \code{earth} image using complex moments.}
	\label{fig:cmplxReconstruct}
\end{figure}

Figures \ref{fig:orthDiscReconstruct}, \ref{fig:orthContReconstruct}, and \ref{fig:cmplxReconstruct} show the \code{earth} image reconstructed using the different moment types at different orders. 

\begin{figure}[H]
	\centering
	\includegraphics[width=\textwidth]{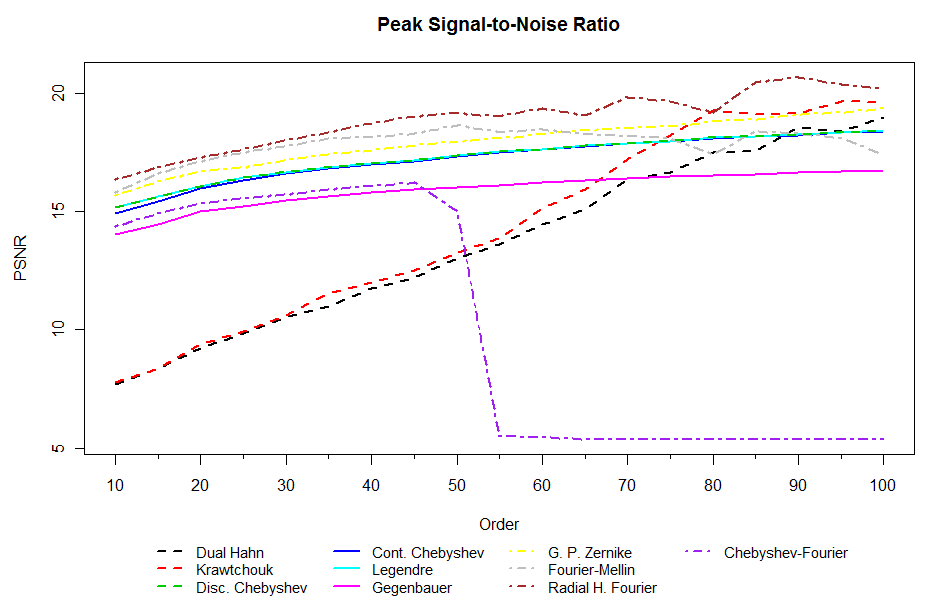}
	\caption{Peak signal-to-noise ratio of the original \code{earth} image to reconstructed images from different types of moments at different orders.}
	\label{fig:allError2}
\end{figure}

Among orthogonal moments at higher orders, Krawtchouk moments yield the best reconstruction. Gegenbauer moments perform better at lower orders, but the quality does not increase significantly as the order increases. Legendre and discrete Chebyshev moments perform similarly, as do dual Hahn and Krawtchouk moments. All four complex moment types perform similarly at all orders except for Chebyshev-Fourier, with which the reconstructed image disappears at higher orders. Among the complex moment types, generalized pseudo-Zernike provides the best reconstruction. It should be emphasized that the calculation of complex moments takes much longer than the calculation of orthogonal moments. Figure \ref{fig:allError2} and table \ref{tab:allErrorTable} show the reconstruction error using different moment types at different orders. 

\begin{table}[H]\footnotesize
	\centering
	\begin{tabular}{lccccc}
		\toprule
		&& \multicolumn{4}{c}{Order}\\
		\cmidrule{3-6}
		& Parameters & 10 & 25 & 50 & 75 \\
		\midrule
		Discrete Chebyshev 	& 	&15.16548		&16.43295		&17.35355		&17.98789\\
		Krawtchouk 			& 0.5	&7.781064		&9.916283		&13.269029		&18.197378\\
		Dual Hahn 		&160, 64	&7.697154		&9.849374		&13.030492		&16.654635\\
		Gegenbauer 			& 2	&14.04270		&15.22617		&16.02905		&16.46257\\
		Continuous Chebyshev 	& 	&14.89885		&16.32907		&17.33970		&17.96409\\
		Legendre 			& 	&15.16554		&16.43145		&17.35424		&17.97658\\
		\\
		Generalized Pseudo-Zernike& 1 	&15.66411		&16.84878		&17.94048		&18.61005\\
		Fourier-Mellin 		& 	&15.81839		&17.47692		&18.64560		&18.09396\\
		Radial Harmonic Fourier 	& 	&16.36668		&17.60459		&19.14355		&19.63776\\
		Chebyshev - Fourier 	& 	&14.359861		&15.559318		&14.982753		&5.385998\\
		\bottomrule
	\end{tabular}
	\caption{Peak signal-to-noise ratio computed at some orders of different moment types.}
	\label{tab:allErrorTable}
\end{table}

\section{Computation time}
To evaluate the running time of various moments, a list of 20 images of size $64 \times 64$, $256 \times 256$, $512 \times 512$, and $1024 \times 1024$ pixels were analyzed. The computation times for calculating moments and reconstructing the images are displayed in Table \ref{tab:runningTable}. Four sample images provided with the \pkg{IM} package, \code{lena}, \code{mandril}, \code{livingroom}, and \code{pirates} were each duplicated five times to create a list of 20 images. Generalized pseudo-Zernike, Legendre, and discrete Chebyshev types are compared. The calculation times within each moment type group (continuous orthogonal, discrete orthogonal, and continuous complex moment types) are similar. The computation time for discrete Chebyshev and Legendre moments is significantly faster because they are computed using matrix algebra, while generalized pseudo-Zernike moments must be computed iteratively. This is also true for all types of complex moments computed over the polar coordinates of pixels.

\begin{table}[H]
	\centering
	\scalebox{0.85}{
		\begin{tabular}{ccclllllll}
			&&& \multicolumn{7}{c}{Computation Time in Seconds}\\
			\toprule
			&&& \multicolumn{3}{c}{Moment Calculation}  & \multicolumn{3}{c}{Reconstruction} \\
			\cmidrule{4-6}  \cmidrule{8-10}
			Size & Order & & \pbox{5cm}{GPZM \\ ($\alpha=0$)} & \pbox{5cm}{Discrete \\ Chebyshev} & Legendre & & \pbox{5cm}{GPZM \\ ($\alpha=0$)} & \pbox{5cm}{Discrete \\ Chebyshev} & Legendre \\ 
			\cmidrule{1-2} \cmidrule{4-6} \cmidrule{8-10} \\
			
			& 10 & & 0.158 & 0.059 & 0.057 & & 0.248 & 0.020 & 0.021\\
			$64 \times 64$ 	& 50 & & 1.702 & 0.060 & 0.044 & & 4.329 & 0.042 & 0.040\\
			&100 & & 5.485 & NA & NA & &16.628 & NA & NA\\
			\\
			&10 & & 1.779 & 0.097 & 0.093 & & 3.692 & 0.040 & 0.048\\
			$256 \times 256$ &50 & & 1.117(min) & 0.100 & 0.097 & & 1.158(min) & 0.053 & 0.068\\
			&100 & & 4.278(min) & 0.121 & 0.106 & & 4.420(min) & 0.086 & 0.097\\
			\\
			& 10 & & 14.514 & 0.278 & 0.235 & & 14.687 & 0.139 & 0.103\\
			$512\times 512$  & 50 & & 4.467(min) & 0.279 & 0.256 & & 4.497(min) & 0.168 & 0.188\\
			&100 & & 17.231(min) & 0.284 & 0.272 & & 17.813(min) & 0.179 & 0.210\\
			\\
			& 10 & & 58.432 & 0.887 & 0.879 & & 58.959 & 0.511 & 0.542\\
			$1024 \times 1024$ & 50 & & 24.244(min) & 0.935 & 0.923 & & 18.489(min) & 0.564 & 0.587\\
			& 100 & & 1.219(hr) & 1.001 & 0.967 & & 1.353(hr) & 0.706 & 0.665\\
			\\
			\bottomrule
	\end{tabular}}
	\caption{Computation time for generalized pseudo-Zernike, discrete Chebyshev, and Legendre moments and image reconstruction.}
	\label{tab:runningTable}
\end{table}

\section{Conclusions}
Moments and moment invariants are useful and established tools for image processing, recognition, and classification, capturing shapes and textures present in images in varying detail. Many types of polynomials exist for calculating moments, and each has its advantages and disadvantages. While not rotation invariant, real orthogonal moments perform reconstructions superior to those of complex moments. However, one should still be aware that while complex moments may not perform well in image reconstruction, they can be easily made rotation invariant and, therefore, may perform better as features in various classification applications. If speed is essential to the application, using a polar transform in conjunction with a discrete orthogonal moment type may be a more efficient method. The type of moment chosen may depend on many factors, such as how high order is needed for the application, whether reconstruction or classification is a priority, what type of image is being processed, and whether there are restrictions on processing power and/or running time. 

The \pkg{IM} package provides tools for the computation of image moments. It also provides a polar transform that approximates rotational invariance for real orthogonal moment types, which are computed quickly and easily. There are several ways to accomplish the same tasks using the \pkg{IM} package, providing the user with flexibility and access to different moment calculation and image reconstruction stages.

\section*{Acknowledgment}
This research was supported through the National Institute of Allergy and Infectious Diseases (NIAID), grant number: 5R21AI085531-02 and the Agricultural Research Service of the US Department of Agriculture, grant number: 1935-42000-035.

\nocite{*}

\printbibliography







\end{document}